%% file: main.tex
\documentclass[runningheads]{llncs}

\usepackage{eccv}

\usepackage{eccvabbrv}

\usepackage{graphicx}
\usepackage{booktabs}

\usepackage[accsupp]{axessibility}  

\usepackage{hyperref}

\usepackage{orcidlink}

\usepackage{adjustbox}
\usepackage{booktabs}
\usepackage{multirow}
\usepackage{float}
\usepackage{colortbl}
\usepackage{siunitx}
\usepackage{arydshln}
\usepackage{amssymb} 
\usepackage{tikz}
\usepackage{float}
\usepackage{booktabs}
\usepackage{makecell}
\usepackage{multirow}
\usepackage{amssymb}
\usepackage{graphicx}
\usepackage{enumitem}
\usepackage[most]{tcolorbox}
\usepackage[table,dvipsnames]
{xcolor}
\usepackage{titletoc}
\newcommand{\authcount}[1]{}
\begin{document}

\title{VKnowU: Evaluating Visual Knowledge Understanding in Multimodal LLMs}

\titlerunning{VKnowU}

\author{Tianxiang Jiang\inst{1,2}\orcidlink{0009-0001-3957-9748} \and
Sheng Xia\inst{3,4}\orcidlink{0009-0005-8889-3927} \and
Yicheng Xu\inst{2}\orcidlink{0000-0003-2975-1206} \and
Linquan Wu\inst{5}\orcidlink{0009-0000-6594-625X} \and \\
Xiangyu Zeng\inst{3,2}\orcidlink{0000-0001-6956-5040} \and
Limin Wang\inst{3,2}\orcidlink{0000-0002-3674-7718} \and
Yu Qiao\inst{2}\orcidlink{0000-0002-1889-2567} \and
Yi Wang\inst{2}\thanks{Corresponding author.}\orcidlink{0000-0001-9134-1203}}

\authorrunning{T.~Jiang et al.}

\institute{
$ ^{1}$University of Science and Technology of China \quad 
$ ^{2}$Shanghai AI Laboratory \\
$ ^{3}$Nanjing University 
$ ^{4}$Shanghai Innovation Institute 
$ ^{5}$City University of Hong Kong
\\
\email{
\url{https://github.com/OpenGVLab/VKnowU}
}
}


\def\MODEL{VideoKnow+}
\def\DATA{VKnowQA}
\def\BenchName{VKnowU}
\def\BENCH{VKnowU}
\definecolor{darkblue}{RGB}{0, 0, 139}
\definecolor{lightblue}{RGB}{173, 216, 230}
\definecolor{myblue}{HTML}{E1ECF1}
\definecolor{myyellow}{HTML}{FDEAD2}
\definecolor{blue2}{HTML}{7497BE}
\definecolor{yellow2}{HTML}{FBC98C}
\definecolor{mygreen}{HTML}{C7E6D3}
\definecolor{highlightgreen}{HTML}{C7E6D3}

\maketitle
\input{sec/0_abstract}   
\input{sec/1_intro}
\input{sec/2_related}
\input{sec/3_DarkBench}
\input{sec/4_Method}
\input{sec/5_Exp}
\input{sec/6_Conclusion}

\clearpage

%
%
\bibliographystyle{splncs04}
\bibliography{main}

\input{sec/Supplementary_Material}

\end{document}

%% file: sec/0_abstract.tex
\begin{figure}[ht]
    \centering
\includegraphics[width=.99\textwidth]{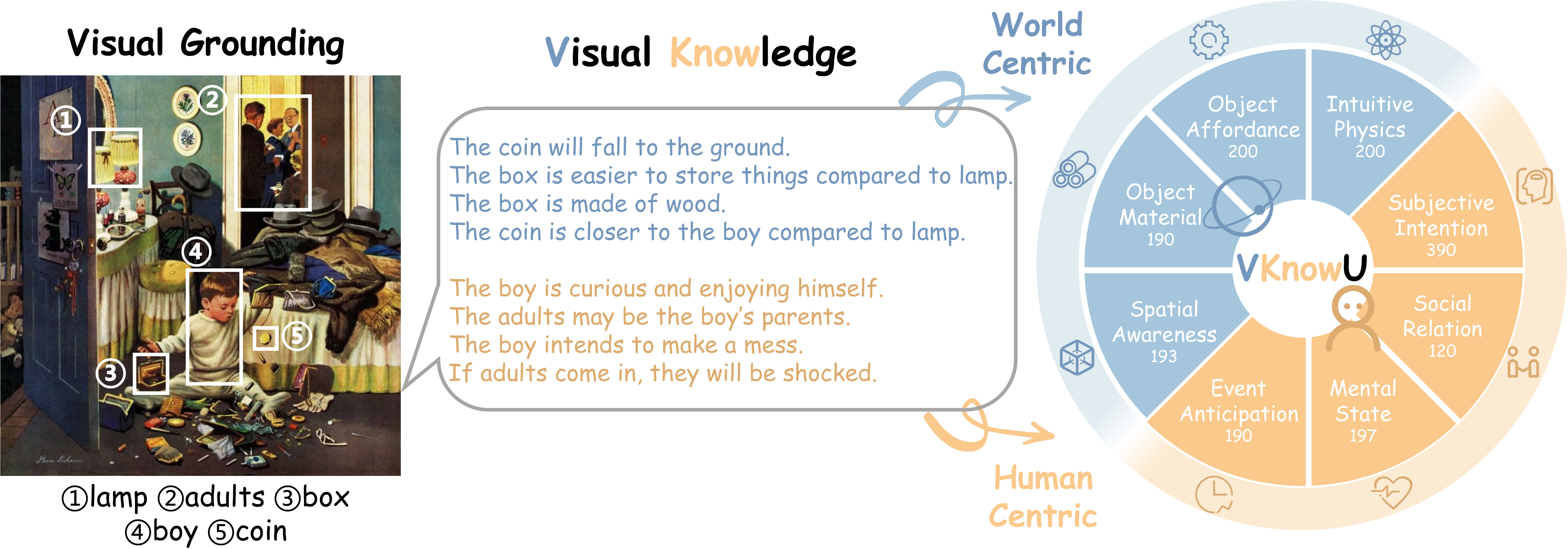}
        \captionof{figure}{\textbf{\BENCH} systematically evaluates visual knowledge understanding of MLLMs across world-centric and human-centric tasks. It marks a shift from mere seeing to true understanding of our physical and social worlds.}%
        \label{fig:from_grounding_to_knowledge}%
\end{figure}

\begin{abstract}
While Multimodal Large Language Models (MLLMs) have become adept at recognizing objects, they often lack the intuitive, human-like understanding of the world's underlying physical and social principles. This high-level vision-grounded semantics, which we term \textbf{visual knowledge}, forms a bridge between perception and reasoning, yet remains an underexplored area in current MLLMs.
To systematically evaluate this capability, we present \textbf{VKnowU}, a comprehensive benchmark featuring 1,680 questions in 1,249 videos, covering 8 core types of visual knowledge spanning both \textbf{world-centric} (e.g., intuitive physics) and \textbf{human-centric} (e.g., subjective intentions). Evaluation of 28 SOTA MLLMs reveals that leading models still fall short of human performance, with particularly notable gaps in the world-centric.
To bridge this gap, we introduce a new dataset, \textbf{VKnowQA}, and \textbf{VideoKnow+}, a baseline model that explicitly incorporates visual knowledge into MLLMs. VideoKnow+ follows a structured \textit{See–Think–Answer} paradigm and adopts reinforcement learning with visual knowledge reward, achieving a +3.7\% improvement on VKnowU and consistent gains on MVBench (+5.4\%), Video-MME (+7.0\%), and MMVU (+5.7\%).
Our work highlights visual knowledge as a missing cornerstone for developing more generalizable MLLMs that can not only see but also truly understand our worlds.
\keywords{Visual Knowledge \and Multimodal Large Language Models}
\end{abstract}

%% file: sec/1_intro.tex
\section{Introduction}
Humans possess an intuitive understanding of the world, effortlessly predicting the trajectory of a bouncing ball or inferring the fragility of a glass from a single glance. This ability, referred to as visual knowledge~\cite{ zhu2020dark, wang2025visual}, represents an intermediate cognitive layer that internalizes the principles that govern both our physical and social worlds. For Multimodal Large Language Models (MLLMs), which frame predictions as probabilities conditioned on both visual evidence and world knowledge, this layer is crucial. By grounding reasoning in rich, visually-derived context, visual knowledge helps reduce over-reliance on brittle language priors and mitigates hallucinations. Enhancing this ability is fundamental to advancing MLLMs' performance on complex tasks such as compositional problem-solving~\cite{yue2024mmmumassivemultidisciplinemultimodal} and multi-hop reasoning~\cite{cheng2025videoholmesmllmthinklike}, towards human-like AI~\cite{lee2024towards,liu2025generative,chi2026driver}.

Despite its importance, this critical dimension of visual understanding remains largely unexplored in the context of MLLMs. Existing research has primarily focused on enhancing fine-grained perception through techniques such as leveraging multi-scale features~\cite{jiang2023clip, liu2024sphinx}, aligning vision and text with detailed descriptions~\cite{zhang2024long,wang2024videoclip}, or processing high-resolution inputs~\cite{liu2024improved,wang2025make}. While these approaches improve an MLLM’s ability to recognize concrete objects and attributes, they contribute little to instilling the embedded, intuitive knowledge necessary for deeper reasoning.

In this paper, we first introduce \BENCH\ to quantitatively measure this gap. \BENCH\ is a comprehensive video benchmark designed to evaluate the visual knowledge of MLLMs. Spanning both \textbf{\textit{world-centric}} (Intuitive Physics, Object Affordances, Object Materials, Spatial Awareness) and \textbf{\textit{human-centric}} (Event Anticipation, Mental States, Social Relations, Subjective Intention) domains, \BENCH\ comprises 1,249 videos and 1,680 multiple-choice questions across eight dimensions. Through rigorous data curation, we isolate visual knowledge from confounding cues like audio or language biases, creating a clean and challenging testbed. Our evaluation shows that even the most advanced MLLMs fall short of human performance with an overall gap of \textbf{16.7\%}, which is particularly pronounced in \textit{world-centric} tasks: on Intuitive Physics and Spatial Awareness, model barely exceeds random guessing, trailing human performance by \textbf{37.0\%} and \textbf{32.7\%}, respectively. These findings highlight the limitation in current MLLMs’ ability to comprehend visual knowledge.

Having quantified the problem, we then demonstrate that visual knowledge is learnable. We propose \MODEL, a baseline model designed to explicitly integrate visual knowledge. \MODEL\ employs a structured \textit{See--Think--Answer} reasoning format and is trained with reinforcement learning guided by a dedicated visual knowledge reward signal. We also introduce \DATA, a new large-scale video corpus containing diverse instances of visual knowledge. When trained on \DATA, \MODEL\ achieves a significant \textbf{3.7\%} improvement on \BENCH\ and shows strong generalization to other video understanding benchmarks, including MVBench~(\textbf{+5.4\%})~\cite{li2024mvbench}, Video-MME~(\textbf{+7.0\%})~\cite{fu2025video} and MMVU~(\textbf{+5.7\%})~\cite{zhao2025mmvu}. These results underscore the pivotal role of explicit visual knowledge in advancing multimodal learning.


In summary, our contributions are as follows: (1) We formalize the concept of \textbf{visual knowledge} for MLLMs, highlighting the critical gap between perceptual accuracy and cognitive reasoning in current models.
(2) We introduce \textbf{\BENCH}, a comprehensive video benchmark covering eight dimensions of visual knowledge, carefully curated to minimize confounding cues and provide a challenging testbed. (3) We release \textbf{\DATA}, a large-scale video corpus rich in visual knowledge, and propose \textbf{\MODEL}, a baseline model that explicitly integrates visual knowledge using a structured \textit{See--Think--Answer} format and reinforcement learning with a visual knowledge reward. 


%% file: sec/2_related.tex
\section{Related Work}

\paragraph{\textbf{Multimodal Large Language Models.}}
The rapid rise of Large Language Models (LLMs) has spurred the development of Multimodal LLMs (MLLMs), which integrate visual perception with language reasoning to bridge cross-modal semantic gaps. Modern MLLMs \cite{liu2023visual,dai2023instructblip,ye2023mplug,wang2024internvideo2scalingfoundationmodels,zeng2024timesuite,bai2025qwen2,zhu2025internvl3,yan2026internvideo3} typically employ lightweight alignment modules that map visual features from encoders like CLIP \cite{radford2021learning} or SigLIP \cite{zhai2023sigmoid} into the embedding space of open-source LLMs \cite{grattafiori2024llama3herdmodels, li2025vista}. Fine-tuned on image-text pairs, these models rival or surpass proprietary systems~\cite{hurst2024gpt,comanici2025gemini} on multimodal benchmarks.
Recently, inspired by DeepSeek-R1 \cite{guo2025deepseek}, RL post-training has been adapted to MLLMs via advanced reward design for vision reasoning \cite{huang2025vision,yang2025r1,xia2025visionary,li2025self,wang2025monosr,wu2026lavit, lin2026mmfinereason}, and extended to video domains \cite{feng2025video,li2025videochat,wang2025videorft, zeng2026video, jiang2026imagine, guan2026video, xia2026serlearninggroundvideo, li2026keeping}, marking a shift toward reasoning in dynamic multimodal contexts.

\paragraph{\textbf{Benchmarks for Multimodal LLMs.}}
MLLM evaluation has rapidly evolved alongside its capabilities. Early benchmarks focused on visual perception, such as VQA \cite{antol2015vqa}, captioning \cite{lin2015microsoftcococommonobjects}, and OCR \cite{singh2019towards}. Later works \cite{fu2025mmecomprehensiveevaluationbenchmark, liu2023visual, li2023seedbenchbenchmarkingmultimodalllms, fu2024blink, song2024cognitive} decoupled perception from reasoning in static images, while POPE \cite{li2023evaluatingobjecthallucinationlarge} exposed hallucinations via adversarial questions. In video, the evaluation shifts to temporal reasoning, advancing from single-scene \cite{goyal2017something, xiao2021next} to long-context scenarios \cite{li2024mvbench, liu2024tempcompass, shi2026river}. Recent works emphasize multimodal reasoning \cite{yu2025vrbench, liu2025videoreasonbench} and knowledge-intensive tasks \cite{zhang2024humaneval, zhao2025mmvu, xu2025expvid}, integrating visual evidence with expert-level knowledge to test real-world applicability. In contrast, the questions in \BENCH\ focus on visual knowledge rooted in human psychology and are deliberately formulated to be straightforward, eliminating the need for domain-specific knowledge.

\paragraph{\textbf{Visual Knowledge.}}
David Marr’s classical formulation of computer vision~\cite{marr2010vision} focuses on identifying \textit{what} and \textit{where}, but human intelligence encompasses much more. Zhu \etal \cite{zhu2020dark} proposed FPICU as the \textit{dark matter} of vision, like dark matter in the universe, invisible in pixels, yet essential for visual understanding. This insight has also been introduced by visual knowledge~\cite{pan2019visual, wang2025visual,zhou2026worldvqa}, which not only emerges from perception but also enables visual memory and mental simulation, forming the basis of how we interpret the world. VCR~\cite{li2022representation} introduced the visual commonsense reasoning, yet its scope is limited in human activities. Recently, Li~\etal~\cite{li2024core} uncovered the core knowledge deficits in MLLMs where they consistently underperform on low-level abilities relative to high-level ones, highlighting a persistent disconnect between seeing and truly understanding.

%% file: sec/3_DarkBench.tex
\section{\BenchName}
We introduce \BENCH, a comprehensive video benchmark designed to evaluate the visual knowledge of MLLMs across eight dimensions. Here we first formally describe the semantic scope of visual knowledge, followed by an overview of each task included in \BENCH. We then provide a detailed account of the dataset construction and the QA reformulation pipeline. Finally, we evaluate a range of state-of-the-art MLLMs on \BENCH\ and analyze their performance.    

\subsection{Visual Knowledge}

Visual knowledge extends far beyond perception or the grounding of concrete objects, encompassing a set of abstract and transferable principles rooted in cognitive psychology. These principles shape how we interpret our surroundings, engage in reasoning, and act upon the world. Generally, humans intuitively grasp the underlying patterns of real-world phenomena as self-evident truths, whereas machines cannot directly uncover them merely by recognizing pixels or physical entities. 
For example, when a moving sphere rises instead of falling under gravity, we immediately recognize it as a violation of physical laws;
when seeing transparent, colorless crystals on a table, we readily identify them as glass and infer their fragility relative to metal;
and when observing someone approach a door, we naturally infer their intention to open it and leave.
While these judgments need observable, pixel-level entities, they rely even more heavily on the visual knowledge. When encountering such scenarios, humans often arrive at an immediate and intuitive understanding, akin to a form of visual commonsense~\cite{zellers2019recognition} or a visual conditioned reflex~\cite{andrew2001conditioned}, \textbf{without} any expert-level domain knowledge.

\begin{figure}[t]
    \centering
    \includegraphics[width=1.\linewidth]{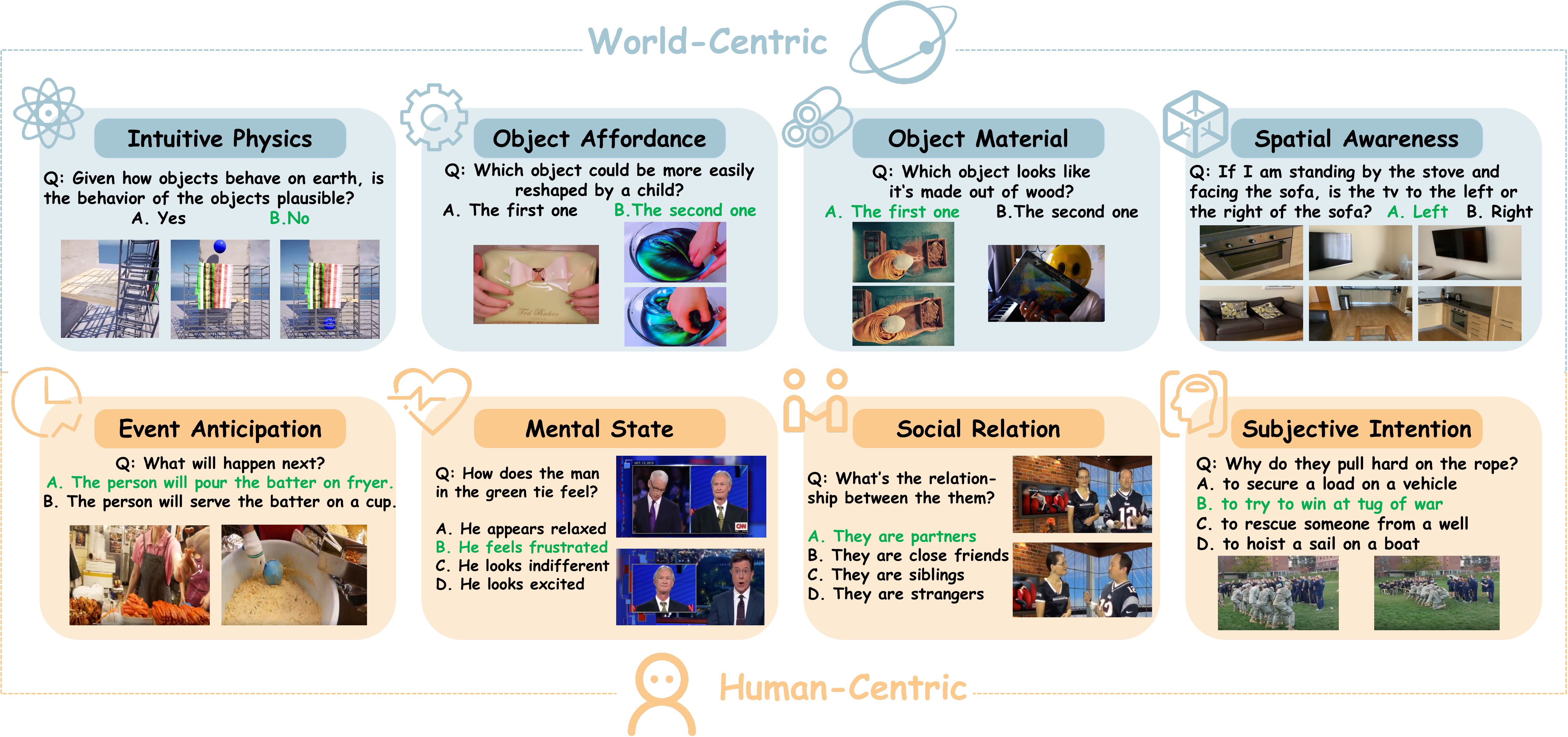}
    \caption{\textbf{Overview of the \BenchName.} Representative video frames and QA from eight tasks across \textit{World-Centric} and \textit{Human-Centric}. Correct answers are shown in green.}
    \label{fig:VKBench}
\end{figure}

\subsection{Task Definition}
\label{sec:task definition}

Based on the meaning of visual knowledge, we concretize its domain by giving its eight most related tasks in {\BENCH}, involving both physical reality (\textbf{\textit{world-centric}}) and psychologically grounded human intentions (\textbf{\textit{human-centric}}). 
The overview of {\BENCH} is shown in Figure \ref{fig:VKBench}.

\paragraph{\textbf{World-Centric.}} 
The \textit{world-centric} categories assess MLLMs' capacity to comprehend the objective physical structure and law of the environment. They are grounded in physics, material properties, and spatial relationships, and are largely independent of human psychological or cultural context. Specifically, they include four core competencies:
(1) \textbf{Intuitive Physics}: Judge the physical plausibility of dynamic events, following principles such as object permanence, immutability, continuity, gravity and so on.
(2) \textbf{Object Affordance}: Determine potential functional uses based on objects' perceptual and structural properties. 
(3) \textbf{Object Material}: Identify the material composition of objects by leveraging either observable intrinsic properties or behavioral cues revealed through human-object interactions.
(4) \textbf{Spatial Awareness}: Infer relative spatial relationships, such as positions, directions, and navigation within an environment.

\paragraph{\textbf{Human-Centric.}} 
The \textit{human-centric} categories evaluate human agents, inferred from visuals such as facial expressions, body postures, gaze, motion trajectories, and social configurations. Contrary to physical knowledge, this is culturally nuanced and psychologically layered. It demands that models reconstruct the invisible mental and social states from visible actions or activities. Core competencies include:
(5) \textbf{Event Anticipation}: Infer the most probable subsequent events based on existing video clues and social norms.
(6) \textbf{Mental State}: Infer internal psychological states of individuals, as well as the affective atmosphere of the surrounding environment.
(7) \textbf{Social Relation}: Infer interpersonal relationships and social roles in human society.
(8) \textbf{Subjective Intention}: Reconstruct the underlying goals or motivations behind human observed actions.

\subsection{\BENCH\ Construction}
\label{sec:bench Construction}
\paragraph{\textbf{Data Collection.}}
We adopt video as the input modality since many target tasks inherently rely on temporal continuity. Static images miss these dynamic signals and often lead to appearance-based shortcuts, whereas videos preserve the rich and essential visual knowledge needed for faithful evaluation. To construct \BenchName, we first collect a large-scale corpus rich in visual knowledge based on multiple datasets (IntPhys 2~\cite{bordes2025intphys}, PACS~\cite{yu2022pacs}, VSI-Bench~\cite{yang2025thinkingspacemultimodallarge}, VLEP~\cite{lei2020more}, Social-IQ 2.0~\cite{siq2} and RexTime~\cite{chen2024rextime}).
Rather than directly reusing original QA annotations, we treat them as \emph{semantic candidates} that will be subsequently reformulated. Each final sample in \BenchName\ is organized as a multiple-choice question designed to probe a specific facet of visual knowledge.

\paragraph{\textbf{QA Reformulation.}} 
\BENCH\ adopts a benchmark reformulation strategy rather than constructing tasks from scratch, to systematically transform existing annotations to isolate visual-knowledge-dependent reasoning. Although the source data provide high-quality annotations, their original design goals are not centered on evaluating visual knowledge. Consequently, many QA instances can be solved through language priors, dataset biases, or auxiliary modalities, which weakens the ability to assess visual knowledge utilization in MLLMs faithfully. This mismatch makes direct reuse of the original annotations unsuitable for our evaluation objective. To address this issue, we align QAs with the visual knowledge taxonomy defined in Sec.~\ref{sec:task definition}. Each retained sample is restructured to explicitly instantiate a predefined visual knowledge competency. This design follows the \textit{reuse-and-refine} philosophy adopted in MVBench~\cite{li2024mvbench}.
Guided by the above principle, we design a progressive reformulation pipeline to ensure that each QA instance genuinely requires visual knowledge while minimizing shortcut solutions induced by non-visual cues. QAs related to \textit{world-centric} categories typically contain limited contextual leakage and are therefore less affected. In contrast, \textit{human-centric} categories often exhibit substantial bias: 
\textbf{(1)} some questions rely heavily on audio signals rather than visual evidence; 
\textbf{(2)} others embed linguistic shortcuts that allow MLLMs to answer without observing the video.   
To address these issues, we apply the following reformulation stages (Fig.~\ref{fig:filter_pipeline}):

\begin{figure}[t]
    \centering
    \includegraphics[width=.95\linewidth]{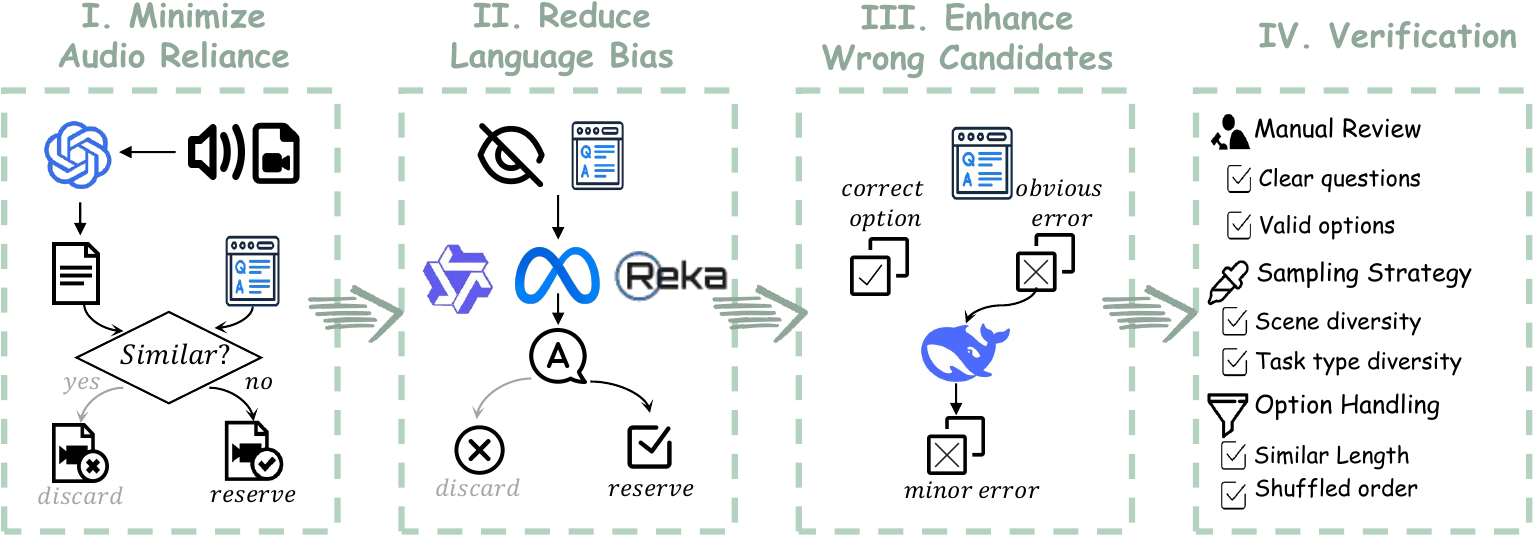}
        \captionof{figure}{\textbf{QA reformulation pipeline} used to construct \BENCH, removing non-visual shortcuts and ensuring each QA requires visual knowledge.}
        \label{fig:filter_pipeline}
\end{figure}

\begin{figure}[t]
    \centering
    \begin{minipage}{0.45\linewidth}
        \centering
\includegraphics[width=\linewidth]{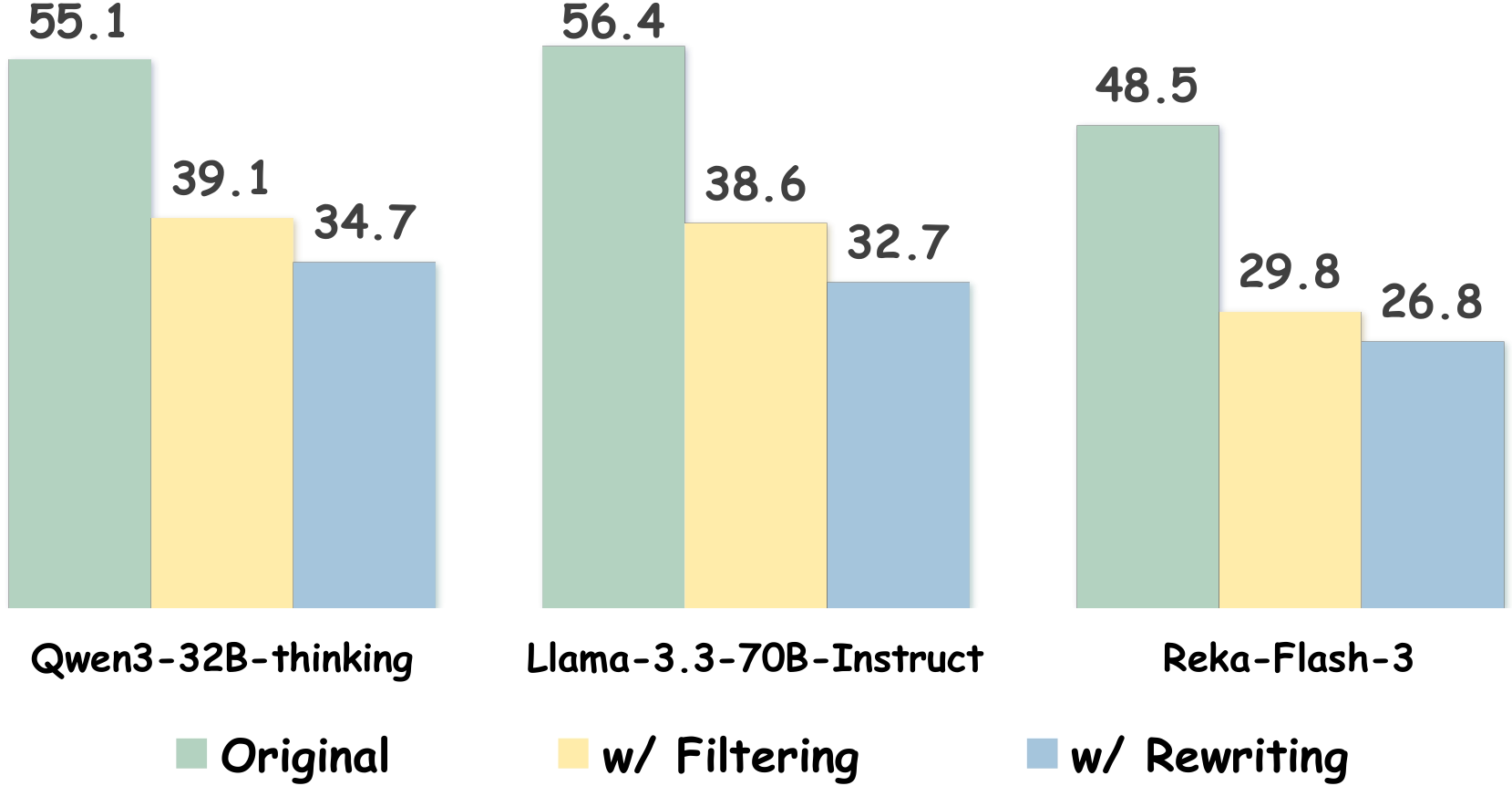}
        \caption{Text-only accuracy downgrades on \BENCH\ after QA reformulation.}
        \label{fig:acc_of_llm}
    \end{minipage}%
    \hfill
    \begin{minipage}{0.53\linewidth}
        \centering
        \begin{subfigure}[b]{0.49\linewidth}
            \centering
            \includegraphics[width=\linewidth]{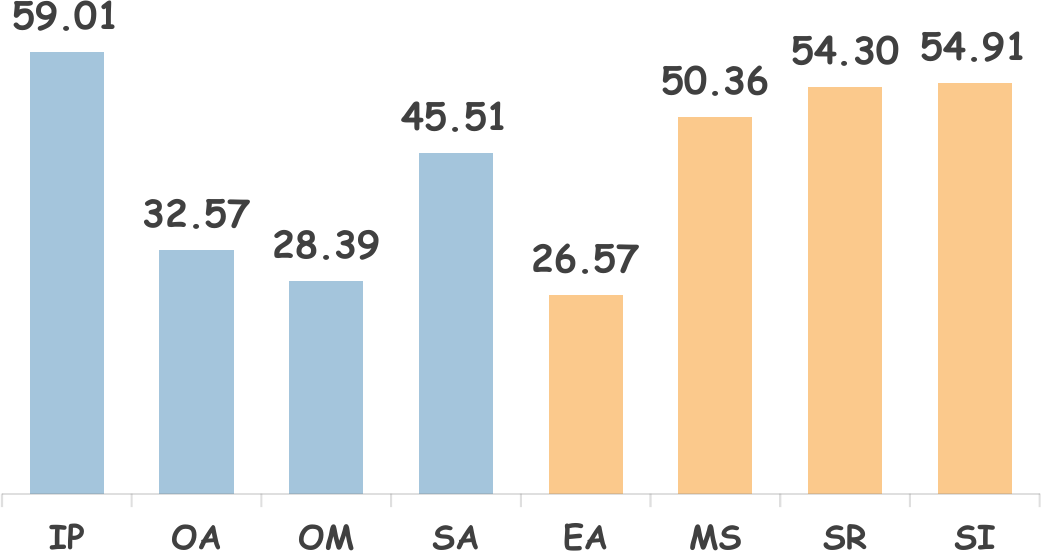}
            \caption{QA word lengths.}
            \label{fig:words_len}
        \end{subfigure}%
        \hfill
        \begin{subfigure}[b]{0.49\linewidth}
            \centering
            \includegraphics[width=\linewidth]{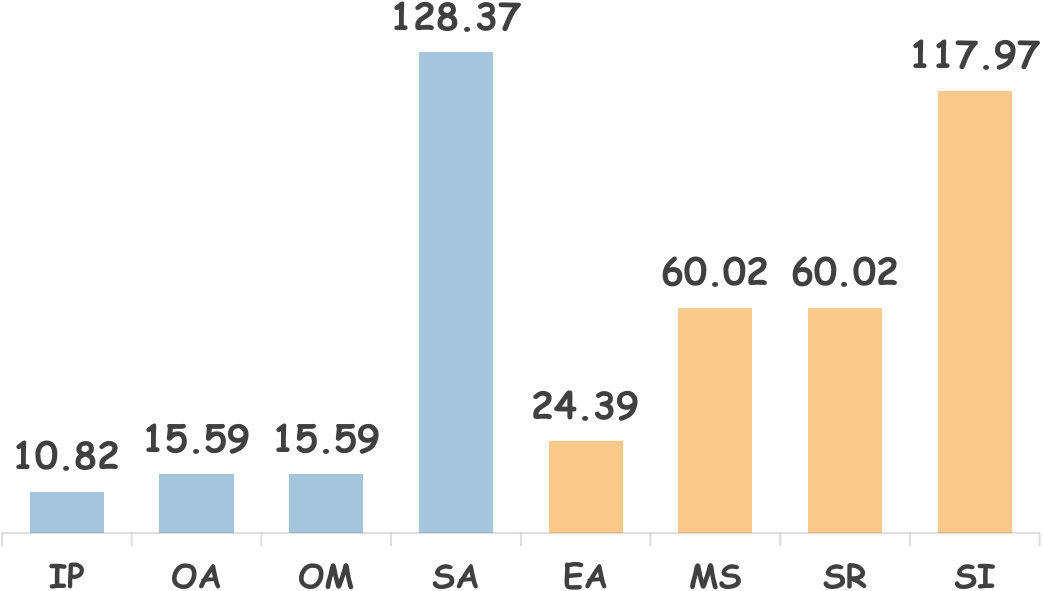}
            \caption{Video duration (s).}
            \label{fig:video_duration}
        \end{subfigure}
        \caption{Average statistics of \BENCH.}
        \label{fig:duration_qa_stats}
    \end{minipage}
\end{figure}

\begin{itemize}[leftmargin=0pt]
    \item \textbf{\uppercase\expandafter{\romannumeral1}. Minimize Audio Reliance.}
    We discard QAs whose answers correlate strongly with audio content. We first transcribe the audio track using Whisper~\cite{radford2022whisper} and compute semantic similarity between subtitles and ground-truth answers using Qwen3-8B-Embedding~\cite{qwen3embedding}. Questions with similarity scores exceeding 0.3 are removed, as they may be solvable through audio cues alone.

    \item \textbf{\uppercase\expandafter{\romannumeral2}. Reduce Language Bias.}
    Inspired by MMMU-Pro~\cite{yue2025mmmuprorobustmultidisciplinemultimodal}, we identify questions solvable without visual inputs. Three text-only LLMs (Reka-Flash-3~\cite{reka}, Qwen3-32B-thinking~\cite{yang2025qwen3} and Llama-3.3-70B-Instruct~\cite{grattafiori2024llama3herdmodels}) are prompted to answer each question under a blind VQA setting. Each model produces ten independent responses. If a model answers correctly more than five times ($\text{Pass@10} \ge 0.5$), the question is deemed linguistically answerable. Questions flagged by at least two models are excluded. As shown in Fig.~\ref{fig:acc_of_llm}, text-only accuracy drops significantly after this stage, indicating improved reliance on visual knowledge.

    \item \textbf{\uppercase\expandafter{\romannumeral3}. Enhance Wrong Candidates.}
    For remaining samples, we reconstruct distractor options using DeepSeek-R1~\cite{guo2025deepseek} while preserving correct answers. The rewritten distractors are designed to be semantically plausible yet subtly incorrect, increasing discriminative difficulty and preventing answer elimination.

    \item \textbf{\MakeUppercase{\romannumeral4}. Human Verification.}
    After shuffling option order to remove positional bias, all questions undergo human verification to ensure clarity, correctness, and alignment with the intended visual knowledge competency.
\end{itemize}

Overall, this process results in \BenchName, comprising 1,249 video clips and 1,680 rigorously curated QA pairs spanning eight dimensions of visual knowledge.

\subsection{Evaluation} 
\label{sec: Evaluation}
 We benchmark a diverse set of multimodal foundation models capable of processing video or multi-image inputs on {\BENCH}, covering both open-source and advanced proprietary MLLMs.
On the open-source side, the evaluated models include VideoLLaMA2~\cite{cheng2024videollama}, mPLUG-Owl3-7B~\cite{ye2024mplug}, MiniCPM-V-2.6 \& 4.5 \cite{yao2024minicpm}, LLaVA-OneVision~\cite{li2024llava}, LLaVA-Video \cite{zhang2024video}, Qwen2.5-VL \cite{bai2025qwen2} \& Qwen3-VL \cite{bai2025qwen3}, InternVL-3.5 \cite{wang2025internvl3}, MiMo-VL-RL \cite{coreteam2025mimovltechnicalreport}, GLM-4.1V and GLM-4.6V \cite{vteam2026glm45vglm41vthinkingversatilemultimodal} . For advanced proprietary models, we include GPT \cite{hurst2024gpt, singh2025openai} and Gemini \cite{comanici2025gemini, blogIntelligenceWith} series. 

\input{tables/bench}

    \paragraph{\textbf{Overall Performance.}}As detailed in Table~\ref{tab:eval_results}, while the SOTA Gemini-3.0-Pro achieves 76.3\%, but still trails human performance (93.0\%) by \textbf{16.7\%}. A distinct competency dichotomy exists between domains. Models demonstrate robustness in \textit{human-centric} tasks (best: 81.9\%), approaching the human baseline (90.6\%), likely benefiting from the prevalence of social narratives in instruction tuning data. In contrast, \textit{world-centric} performance is lacking; the best models in Intuitive Physics and Spatial Awareness lag behind humans by alarming margins of \textbf{37.0\%} and \textbf{32.7\%}, respectively. This deficiency suggests that standard next-token prediction objectives struggle to capture implicit physical dynamics solely from text-aligned supervision, highlighting the potential necessity of integrating world models to ground visual understanding in physical laws.

    \paragraph{\textbf{Proprietary vs. Open-Source MLLMs.}} The evaluation reveals a distinct performance dichotomy. Proprietary models (e.g., Gemini-3.0-Pro) exhibit robust capabilities in \textit{world-centric} tasks, specifically Object Affordance and Object Material, likely benefiting from extensive pre-training on physical world data. Conversely, open-source models demonstrate superior alignment with \textit{human-centric} visual knowledge. Notably, the InternVL3.5 series achieves 81.9\% on \textit{Human-Centric} tasks, surpassing top-tier proprietary counterparts (78.8\%). This asymmetry suggests divergent optimization priorities: while proprietary models emphasize physical grounding, open-source efforts have effectively capitalized on socially diverse supervision to excel in complex human interaction scenarios.
    
    \paragraph{\textbf{Thinking Paradigm.}} Models equipped with reasoning capabilities (e.g., MiMo-VL-RL, GLM-4.1V-Thinking) generally exhibit modest gains over their standard counterparts, particularly in \textit{world-centric} tasks. However, this benefit is not universal; notably, InternVL3.5-38B-Think suffers a performance regression due to severe repetition during reasoning. We hypothesize an \textit{inverted-U} relationship between reasoning length and accuracy: while some deliberation aids in grounding visual evidence, excessive chain-of-thought introduces noise, especially for tasks in \BENCH\ which rely on intuitive, almost reflexive visual knowledge. This aligns with recent findings~\cite{luo2025thinking} that optimal reasoning depth is task-dependent.

\begin{figure*}[t]
    \centering
    \begin{minipage}{0.56\linewidth}
        \centering
        \includegraphics[width=\linewidth]{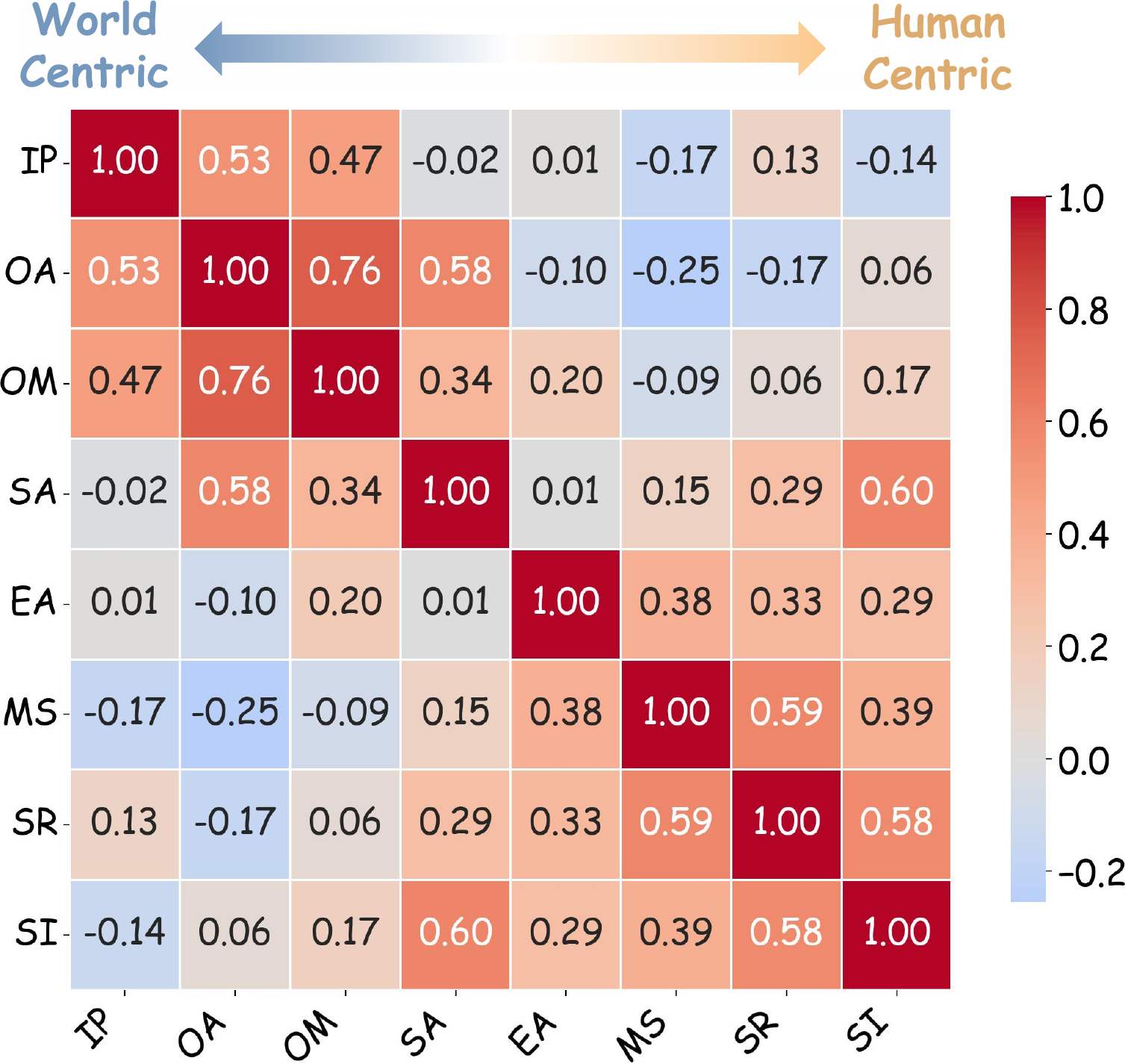}
        \captionof{figure}{\textbf{Pearson correlation} among the 8 visual knowledge tasks. Two clear clusters emerge: \textbf{world-centric} and \textbf{human-centric}.} 
        \label{fig:pearson}
    \end{minipage}%
    \hfill
    \begin{minipage}{0.4\linewidth}
        \begin{minipage}{\linewidth}
            \centering
            \includegraphics[width=\linewidth]{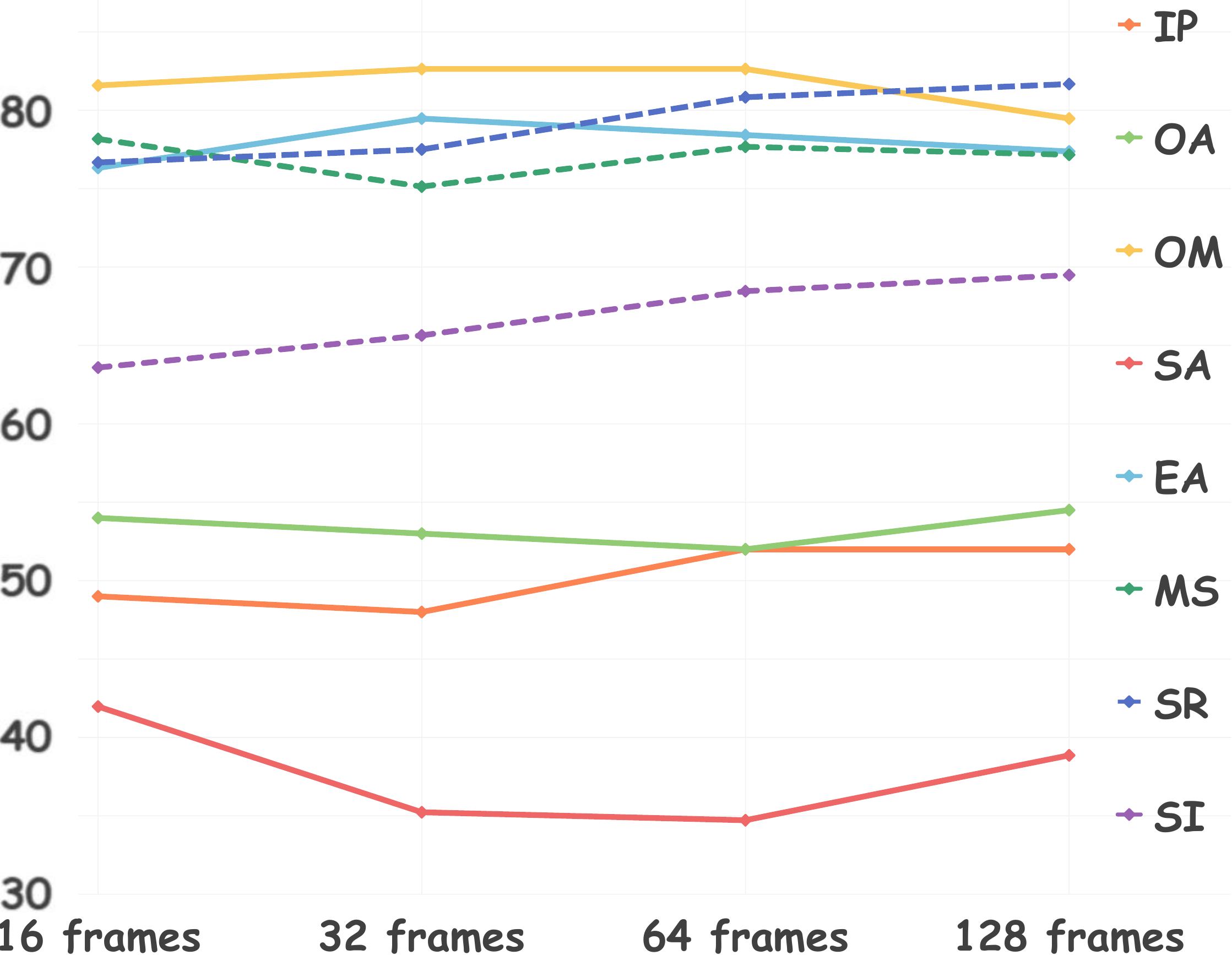}
            \captionof{figure}{\textbf{Frames sensitivity} for each task of \BENCH.}
            \label{fig:frames}
        \end{minipage}
        
        
        \begin{minipage}{\linewidth}
            \centering
            \captionof{table}{\textbf{FPS sensitivity} for world-centric task of \BENCH.}
            \label{tab:REBUTTAL_fps}
            \resizebox{\linewidth}{!}{
                \begin{tabular}{cccccc}
                \toprule
                \textbf{FPS} & \textbf{WC} & \textbf{IP} & \textbf{OA} & \textbf{OM} & \textbf{SA} \\
                \midrule
                \textbf{1} & 55.17 & 49.50 & 51.50 & 80.53 & 39.90 \\
                \textbf{2} & 55.17 & 48.00 & 51.00 & 82.11 & 40.41 \\
                \textbf{3} & 55.94 & 50.00 & 53.50 & \textbf{82.11} & 38.86 \\
                \textbf{4} & \textbf{56.58} & \textbf{51.50} & \textbf{53.50} & 81.58 & \textbf{40.41} \\
                \bottomrule
                \end{tabular}
            }
        \end{minipage}
    \end{minipage}
\end{figure*}

    \paragraph{\textbf{Correlation between Different Tasks.}} As illustrated in Fig.~\ref{fig:pearson}, the Pearson correlation heatmap reveals a distinct block-diagonal structure. The strong intra-cluster coherence within \textit{world-centric} and \textit{human-centric} domains, contrasted with minimal inter-cluster correlations, empirically validates \BENCH's taxonomic framework. This pattern demonstrates that these two dimensions operate as largely independent cognitive subsystems, confirming the necessity of our dual-track evaluation strategy. By disentangling physical grounding from social reasoning, \BENCH\ provides a more precise diagnostic tool for MLLM capabilities than other monolithic benchmarks~\cite{li2022representation}.

    \paragraph{\textbf{Visual Sampling Sensitivity.}} We analyze the impact of visual sampling using Qwen2.5-VL-7B. As shown in Fig.~\ref{fig:frames}, performance sensitivities vary by domain. \textit{Human-centric} tasks (e.g., SR, SI) consistently benefit from increased frame counts, indicating a reliance on rich, accumulated temporal context. Conversely, \textit{world-centric} tasks (e.g., SA) exhibit non-monotonic trends, suggesting that excessive frames may introduce redundancy or noise rather than clarity. This underscores the need for adaptive keyframe selection akin to human attention. Furthermore, Table~\ref{tab:REBUTTAL_fps} reveals that higher FPS improves \textit{world-centric} accuracy. These results confirm that dense temporal sampling is critical for capturing continuous motion dynamics in physical grounding, whereas sparse sampling often suffices for static semantics.

    \paragraph{\textbf{LLM Scaling}.}
    While scaling LLM (e.g., from 7B to 72B) generally yields gains, it is crucial to determine if this stems from genuine visual understanding or stronger language priors. To isolate the impact of the language component, we conduct a controlled LLaVA-style scaling study \cite{liu2024improved} where only the LLM backbone varies (Qwen3-4B vs. 8B~\cite{yang2025qwen3}) while keeping other components fixed. 
    As shown in Table~\ref{tab:REBUTTAL_LLAVA}, increasing the LLM size drives larger improvements on general benchmarks like Video-MME and MMVU. However, gains on \BENCH\ are notably marginal (+1.3\%). Crucially, performance on IP, SA and EA actually degrades. This counter-intuitive trend strongly suggests that \BENCH\ performance is not primarily limited by language modeling capacity, but rather by visual grounding and visual knowledge acquisition. 
    These findings confirm that \BENCH\ evaluates capabilities beyond pure language priors, positioning visual knowledge as a \textit{bridge} between perception and reasoning.



\begin{table}[t]
\caption{\textbf{LLaVA-style scaling study} with different LLMs (Qwen3-4B vs. 8B): \BENCH\ relies on visual knowledge acquisition rather than pure language priors.}
\label{tab:REBUTTAL_LLAVA}
\centering
\resizebox{\columnwidth}{!}{
\begin{tabular}{c|ccccccccc|c|cc}
\toprule
\textbf{Model} & \textbf{VKnowU} & \textbf{IP} & \textbf{OA} & \textbf{OM} & \textbf{SA} & \textbf{EA} & \textbf{MS} & \textbf{SR} & \textbf{SI} & \textbf{Video-MME} & \textbf{MMVU} \\
\midrule
\textbf{LLaVA-Qwen3-4B} & 59.4 & \textbf{53.00} & 52.00 & 54.21 & \textbf{45.60} & \textbf{76.84} & 70.56 & 74.17 & 56.92 & 49.6 & 54.6 \\
\textbf{LLaVA-Qwen3-8B} & \textbf{60.7} & 51.50 & \textbf{57.50} & \textbf{58.42} & 42.49 & 76.32 & \textbf{72.59} & \textbf{75.00} & \textbf{59.23} & \textbf{53.8} & \textbf{58.7} \\
\textbf{\textit{Gain ($\Delta$)}} & \textcolor{ForestGreen}{+1.3} & \textcolor{red}{-1.50} & \textcolor{ForestGreen}{+5.50} & \textcolor{ForestGreen}{+4.21} & \textcolor{red}{-3.11} & \textcolor{red}{-0.52} & \textcolor{ForestGreen}{+2.03} & \textcolor{ForestGreen}{+0.83} & \textcolor{ForestGreen}{+4.36} & \textbf{\textcolor{blue}{+4.2}} & \textbf{\textcolor{blue}{+4.1}} \\
\bottomrule
\end{tabular}
}
\vspace{-5mm}
\end{table}

%% file: tables/bench.tex
\begin{table}[h!]
\caption{We report evaluation results on \BENCH\ using accuracy (\%). Abbreviations adopted: \textbf{IP} for Intuitive Physics; \textbf{OA} for Object Affordance; \textbf{OM} for Object Material; \textbf{SA} for Spatial Awareness; \textbf{EA} for Event Anticipation; \textbf{MS} for Mental State; \textbf{SR} for Social Relation; \textbf{SI} for Subjective Intention; \textbf{WC} for overall accuracy on World-Centric; \textbf{HC} for Human-Centric; \textbf{}Human Performance are sourced from original annotation or researchers' responses. \colorbox{mygreen}{Green} marks the best results. A clear gap remains between SOTA MLLMs and human performance, especially on world-centric tasks.}
\label{tab:eval_results}
\centering
\resizebox{.85\linewidth}{!}{
\begin{tabular}{l c cccccccccc}
\toprule
\textbf{Models} & \textbf{Overall} & 
\cellcolor{myblue}\textbf{IP} & \cellcolor{myblue}\textbf{OA} & \cellcolor{myblue}\textbf{OM} & \cellcolor{myblue}\textbf{SA} & 
\cellcolor{myblue}\textbf{WC} & 
\cellcolor{myyellow}\textbf{EA} & \cellcolor{myyellow}\textbf{MS} & \cellcolor{myyellow}\textbf{SR} & \cellcolor{myyellow}\textbf{SI} &
\cellcolor{myyellow}\textbf{HC} \\
\midrule
Random Guess & 37.4 & 50.0 & 50.0 & 50.0 & 31.9 & 45.5 & 50.0 & 25.0 & 25.0 & 25.0 & 30.3\\
Human Performance & 93.0 & 97.5& 93.5& 96.8& 95.4& 95.8& 87.9& 88.1& 89.2& 93.6&90.6\\
\midrule
\multicolumn{12}{l}{\textit{Open-Source MLLMs}} \\
VideoLLaMA2-7B~\cite{cheng2024videollama} & 55.7 & 46.0 & 53.0 & 51.1 & 40.9 & 47.8 & 68.4 & 64.5 & 57.5 & 60.5 & 62.7
\\
MiniCPM-V 2.6~\cite{yao2024minicpm} & 62.3 & 53.0 & 51.5 & 76.5 & 36.5 & 54.3 & 74.0 & 66.5 & 75.0 & 66.5 & 69.2 \\
MiniCPM-V 4.5~\cite{yao2024minicpm} & 65.9 & 50.0 & 58.5 & 83.5 & 39.0 & 57.6 & 78.5 & 77.2 & 76.7 & 72.5 & 73.1 \\
mPLUG-Owl3-7B~\cite{ye2024mplug} & 64.2 & 57.0 & 55.0 & 72.6 & 36.3 & 55.2 & 77.9 & 73.1 & 79.2 & 66.7 & 72.1
\\
LLaVA-OV-7B~\cite{li2024llava} & 63.3 & 52.5 & 56.0 & 83.0 & 38.0 &57.2& 73.0 & 72.1 & 75.8 & 62.5 & 68.6\\
LLaVA-OV-72B~\cite{li2024llava} & 66.9 & 56.0 & 60.0 & 83.0 & 41.0 & 59.9 & 76.0 & 71.6 & 83.3 & 69.0 & 73.0\\
LLaVA-Video-7B~\cite{zhang2024video} & 64.7 & 50.5 & 56.5 & 79.0 & 42.0 & 56.9 & 72.5 & 69.5 & 80.0 & 69.5 & 71.5\\
LLaVA-Video-72B~\cite{zhang2024video} & 69.9 & 57.0 & 65.5 & 86.0 & 44.5 &63.1& 78.0 & 72.6 & 84.2 & 73.8 & 75.8\\

Qwen2.5-VL-3B-Instruct~\cite{bai2025qwen2} & 61.1 & 50.0 & 57.0 & 77.9 & 38.3 &55.7& 71.6 & 73.1 & 71.7 & 57.4 & 65.8\\
Qwen2.5-VL-7B-Instruct~\cite{bai2025qwen2} & 64.0 & 48.0 & 53.0 & 82.6 & 35.2 & 54.5 & 79.5 & 75.1 & 77.5 & 65.6  & 72.2\\

Qwen2.5-VL-72B-Instruct~\cite{bai2025qwen2} & 70.1  & 59.0 & 64.5 & 86.3 & 40.9 & 62.6 & 76.8 & \cellcolor{mygreen}{79.7} & 82.5 & 73.3 & 76.7\\

MiMo-VL-7B-RL~\cite{coreteam2025mimovltechnicalreport} & 68.0 & 56.0 & 66.0 & 80.5 & 49.2 & 62.8 & 78.4 & 75.6 & 80.8 & 65.4 & 72.5\\

GLM-4.1V-9B-Thinking~\cite{vteam2026glm45vglm41vthinkingversatilemultimodal} & 67.6 & 54.0 & 66.5 & 80.5 & 47.0 & 61.9 & 72.5 & 73.1 & 80.8 & 69.8 & 72.6\\

GLM-4.6V-Flash~\cite{vteam2026glm45vglm41vthinkingversatilemultimodal} & 68.4 & 56.0 & 70.5 & 82.1 & 57.0 &66.3& 76.8 & 70.0 & 80.8 & 64.1 & 70.3 \\


Qwen3-VL-4B-Instruct~\cite{bai2025qwen3} & 67.6 & 52.0 & 53.0 & 84.2 & 54.9 & 60.8 & 76.8 & 73.6 & 79.2 & 70.3& 73.6\\

Qwen3-VL-8B-Instruct~\cite{bai2025qwen3} & 69.0 & 51.5 & 57.5 & 83.7 & 57.0 & 62.2 & 76.8 & 75.6 & 80.0 & 72.3& 75.0\\

InternVL3.5-8B~\cite{wang2025internvl3} & 66.9 & 50.5 & 58.0 & 78.9 & 44.6 & 57.9 & 75.3 & 75.6 & 75.8 & 73.8 & 74.8\\
InternVL3.5-8B-Think~\cite{wang2025internvl3} & 67.6 & 50.5 & 63.0 & 80.5 & 42.0 & 58.9 & 76.3 & 74.1 & 78.3 & 74.1 & 75.1\\
InternVL3.5-30B-A3B~\cite{wang2025internvl3} & 71.5 & 49.5 & 65.0 & 82.1 & \cellcolor{mygreen}{62.7} & 64.6 & 75.8 & 77.2 & 81.7 & 77.2 & 77.5\\
InternVL3.5-38B~\cite{wang2025internvl3} & 72.7 & 49.0 & 61.0 & 82.6 & 59.1 & 62.7 & 77.9 & 76.1 & 82.5 & \cellcolor{mygreen}{85.4} & 81.4\\
InternVL3.5-38B-Think~\cite{wang2025internvl3} & 71.9 & 49.0 & 66.5 & 79.0 & 54.9 & 62.2 & 80.5 & 76.1 & 80.8 & 82.1 & 80.3\\
InternVL3.5-241B-A28B~\cite{wang2025internvl3} & 74.6 & 52.5 & 67.5 & 85.8 & 60.6 & 66.4 & \cellcolor{mygreen}{81.6} & 77.2 & 84.2 & 83.6 & \cellcolor{mygreen}{81.9} \\

\midrule
\multicolumn{12}{l}{\textit{Proprietary MLLMs}} \\
GPT-4o~\cite{hurst2024gpt} & 65.7 & 55.0 & 74.0 & 84.7 & 42.5 & 64.0 & 74.7 & 63.5 & 65.0 & 65.9 & 67.1\\
GPT-5-mini~\cite{singh2025openai} & 66.7 & 56.5 & 77.0 & 87.9 & 52.3 & 68.3 & 69.5 & 60.9 & 75.8 & 62.3 & 65.3 \\
Gemini-2.5-Flash~\cite{comanici2025gemini} & 68.8 & 56.0 & 80.0 & 90.0 & 51.8 & 69.3 & 76.8 & 65.5 & 74.2 & 63.6 & 68.2\\
Gemini-2.5-Pro~\cite{comanici2025gemini} & 71.1 & 55.0 & 82.0 & 88.4 & 55.4 & 70.1 & 71.6 & 73.1 & 75.8 & 70.3 &72.0\\
Gemini-3.0-Flash~\cite{blogIntelligenceWith} &76.1&\cellcolor{mygreen}{60.5}&84.0&\cellcolor{mygreen}{94.2}&53.4&73.0&73.7&77.7&\cellcolor{mygreen}{85.0}&80.0&78.8\\
Gemini-3.0-Pro~\cite{blogIntelligenceWith}&\cellcolor{mygreen}{76.3}&58.0&\cellcolor{mygreen}{90.0}&91.6&56.0&\cellcolor{mygreen}{73.9}&74.8&73.6&81.7&81.8&78.5 \\

\bottomrule
\end{tabular}
}
\vspace{-5mm}
\end{table}

%% file: sec/4_Method.tex

\section{Embedding Visual Knowledge into MLLMs}
\label{sec:method}
Here, we make initial attempts to explicitly integrate visual knowledge into MLLMs. 
Our core idea is to encourage MLLMs to rely more evidence rich in visual knowledge during the autoregressive generation, so that we can effectively offload the burden from the language prior and reduce hallucinations (Theoretical analysis in Appendix A). To operationalize this principle, we design an explicit reasoning pathway that anchors responses directly in perceived visual evidence, which is implemented via two key components: (1) a structured \textit{See--Think--Answer} output format that enforces perceiving visual knowledge before reasoning, and (2) a visual knowledge reward based on the GRPO paradigm. Together, these form the baseline {\MODEL}. 

\paragraph{\textbf{\textit{See-Think-Answer} Output.}} 
Inspired by \cite{xiao2025advancing, xia2025visionary}, we constrain the \MODEL\ to explicitly generate a self-contained visual description prior to reasoning. This description is designed to encapsulate the rich visual knowledge essential for the task, thereby laying a solid foundation for subsequent reasoning. To enforce strict adherence to the structured \textit{See--Think--Answer} output, format reward $r_f$ is computed via regular expression matching over the model's output. For multiple-choice QA, we define an accuracy reward $r_a$ that equals $1$ if the model's prediction matches the ground-truth answer, and $0$ otherwise.


\paragraph{\textbf{Visual Knowledge Reward.}}
\label{sec:visual knowledge reward}
 We employ an external frozen MLLM as the verifier model to assess whether the correct answer can be directly inferred from the generated visual description. To encourage the model to produce self-contained content rich in visual knowledge, we incorporate a binary visual knowledge reward $r_v \in \{0, 1\}$ into the policy update process, where $r_v = 1$ indicates that the visual content is sufficient for deriving the answer without further reasoning and $r_v = 0$ otherwise. This reward incentivizes the model to focus more on enriching its visual knowledge, the richer the visual knowledge, the less reasoning effort is required from the LLM. The final reward is formulated as:
\begin{equation}
    R_i = r_f + r_a + \lambda\cdot r_v,
\end{equation}
where $\lambda$ is a hyperparameter that controls the visual knowledge reward weight.

\subsection{\DATA: New Visual Knowledge Datasets}
To meet the training requirements of \MODEL, we construct two training datasets, named \textbf{\DATA-30K} and \textbf{\DATA-CS-12K}. Each instance in \DATA\ consists of a question accompanied by multiple-choice options. We carefully perform deduplication to ensure no overlap between \DATA\ and \BENCH, thereby avoiding data leakage and guaranteeing an \textbf{out-of-domain} evaluation setting.

\begin{figure}[t]
    \centering
    \begin{minipage}{0.28\linewidth}
        \centering
        \includegraphics[width=\linewidth]{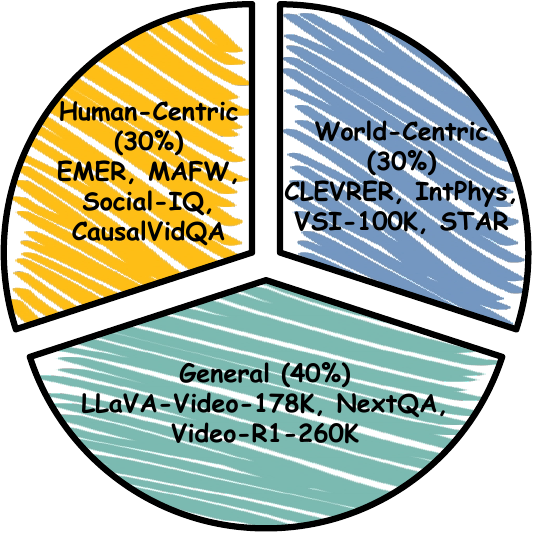}
            \caption{Data composition of \DATA-30K.}
    \end{minipage}%
    \hfill
    \begin{minipage}{0.68\linewidth}
        \centering
        \begin{subfigure}[b]{0.49\linewidth}
            \centering
            \includegraphics[width=\linewidth]{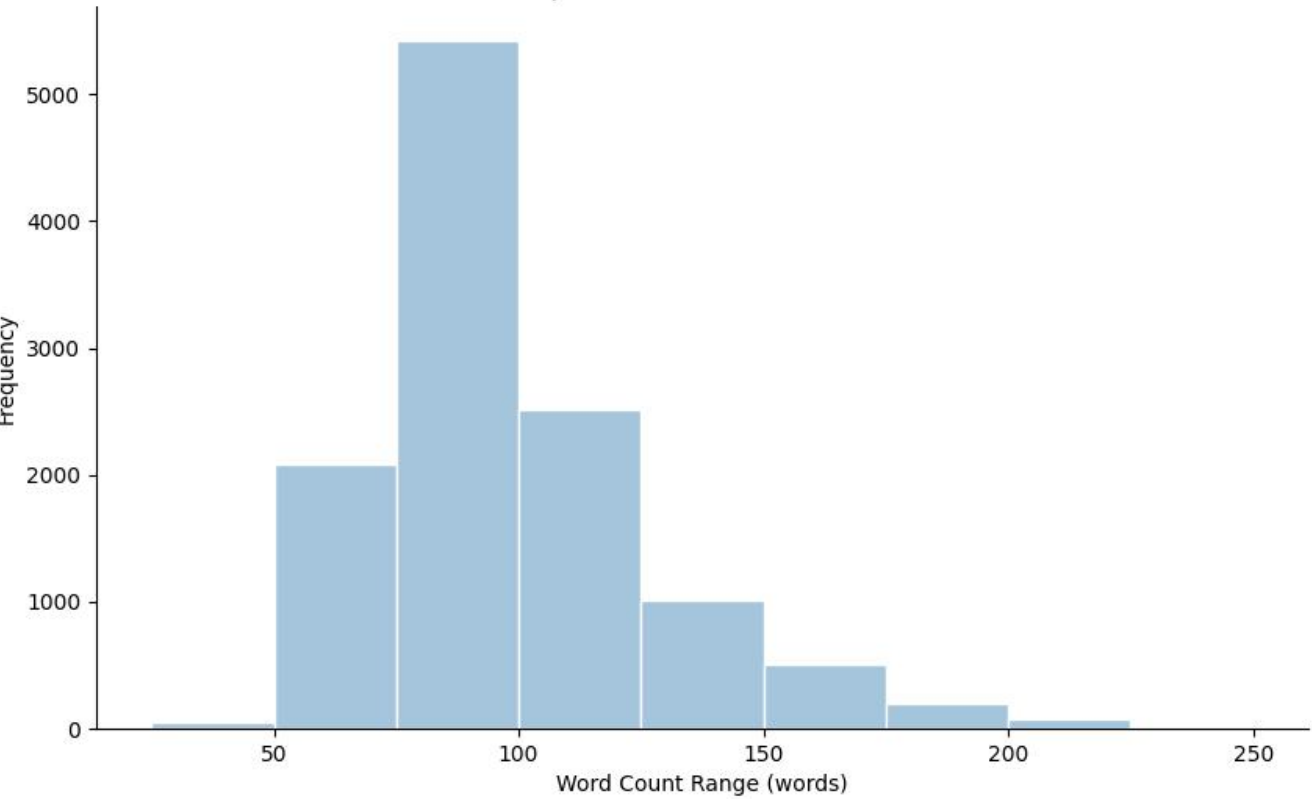}
            \caption{\textit{description}.}
            \label{fig:desc_tokens}
        \end{subfigure}%
        \hfill
        \begin{subfigure}[b]{0.49\linewidth}
            \centering
            \includegraphics[width=\linewidth]{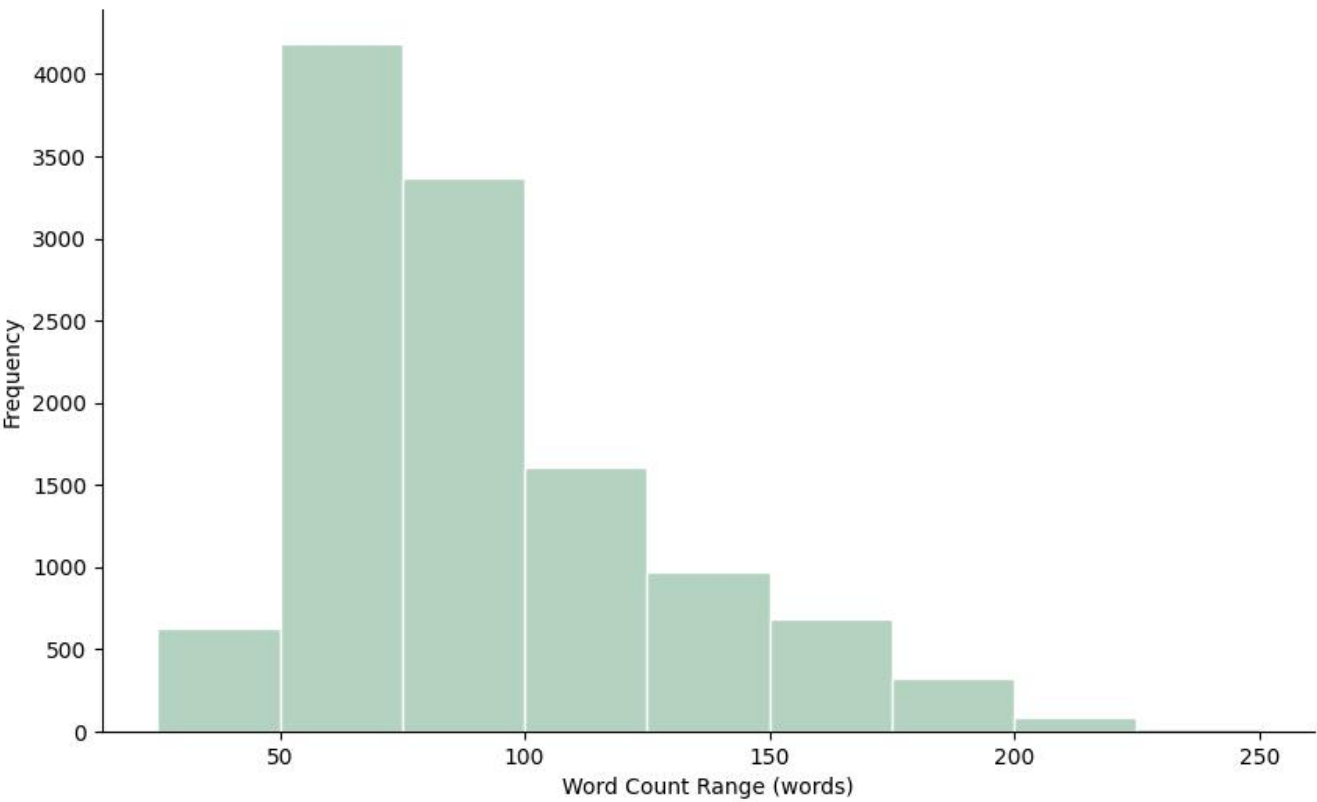}
            \caption{\textit{think}.}
            \label{fig:think_tokens}
        \end{subfigure}
    \caption{Word length distribution of VKnowQA-CS-12K.}
    \end{minipage}
\end{figure}


\paragraph{\textbf{RL Data Collection.}}
Our reinforcement learning dataset comprises approximately 30K samples by augmenting data from open-source VLM datasets, named \DATA-30K. The dataset covers three major domains: general (40\%), world-centric (30\%), and human-centric (30\%). These instances encapsulate diverse and rich visual knowledge across the aforementioned dimensions.

\paragraph{\textbf{Cold Start Data Generation.}}
We also provide data in the \textit{See--Think--Answer} format to facilitate SFT scenarios. We first employed an MLLM (Qwen2.5-VL-7B, see more details in Appendix D) to generate \textit{See-Think-Answer} responses and only retained those cases where the model produced both a correct answer and an output in desired format. Following the similar manner in Sec.~\ref{sec:visual knowledge reward}, we then ensured the model could correctly answer question using only the text-based visual content as a proxy for the visual input, thereby guaranteeing that the reasoning was grounded in explicit visual knowledge. This process resulted in about 12K high-quality QAs, which we denote as \DATA-CS-12K.

\input{tables/main_experiment}

%% file: tables/main_experiment.tex
\begin{table}[t]
\caption{\textbf{Performance across \BENCH\ and other video benchmarks.} 
GRPO-Zero indicates GRPO without SFT like DeepSeek-R1-Zero~\cite{guo2025deepseek}. All results are obtained using 256×28×28 resolutions and max 32 frames. See Appendix E for details.}
\label{tab:main_exp}
\centering
\resizebox{0.95\linewidth}{!}{
    \begin{tabular}{lcccccc}
    \toprule
    \textbf{Models} &  \textbf{\BENCH} & \textbf{MVBench} & \textbf{Video-MME} & \textbf{MMVU} & \textbf{VSI-Bench} &\textbf{Avg.}  \\
    \midrule
    \multicolumn{6}{l}{\textit{Open-Source Models}} \\
    VideoLLaMA2~\cite{cheng2024videollama} &55.7&54.6&47.9&44.8&-&-\\
    mPLUG-Owl3-8B~\cite{ye2024mplug} &64.2&54.5&53.5&-&-&-\\
    LLaVA-OneVision-7B~\cite{li2024llava} &63.3&56.7 &58.2&49.2&34.1& 52.30\\
    \midrule
    \multicolumn{6}{l}{\textit{Recent R1-based Video Methods}} \\
    Video-R1-7B~\cite{feng2025video} &65.3&63.9 &59.3&63.8&30.8&56.62\\
    VideoRFT-7B~\cite{wang2025videorft} &64.6&62.1 &\textbf{59.8}&\textbf{68.5}&35.5&58.10\\
    \midrule
    \multicolumn{6}{l}{\textit{Basemodel: Qwen2.5-VL-7B-Instruct}} \\
    \rowcolor{gray!30}
    Zero-Shot & 64.0 & 59.4 & 52.8 & 61.3 &34.8 & 54.46\\
    \textit{See-Think-Answer} SFT & 63.9&61.0&53.3&63.2&35.1 & 55.30\\
    GRPO-Zero &64.3 & 63.8 & 58.8 & 65.6 &23.6 &55.22\\
    \rowcolor{mygreen}
    \textbf{\MODEL} & \textbf{67.7} & \textbf{64.8} & \textbf{59.8} & 67.0 &\textbf{35.9} &\textbf{59.04} \\
    \bottomrule
    \end{tabular}
}
\vspace{-2mm}
\end{table}

%% file: sec/5_Exp.tex
\section{Experiments}
\label{sec:exp}
\subsection{Main Results}
We adopt Qwen2.5-VL-7B-Instruct \cite{bai2025qwen2} as the base model for validation. To achieve the structured \textit{See--Think--Answer} output format, we first conduct a SFT cold-start phase. This is followed by training with GRPO augmented with the visual knowledge reward $r_v$, to encourage the model to generate responses by explicitly leveraging the visual knowledge it has acquired. This two-stage process yields the final \MODEL. The implementation details are in Appendix E.


As shown in Table~\ref{tab:main_exp}, our model achieves a \textbf{3.7\%} improvement over the baseline on \BENCH, highlighting its superior capacity to comprehend visual knowledge. Moreover, \MODEL\ demonstrates strong generalization across multiple video understanding and reasoning benchmarks, surpassing the baseline by \textbf{5.4\%} on MVBench, \textbf{7.0\%} on Video-MME, \textbf{5.7\%} on MMVU and \textbf{1.1\%} on VSI-Bench, while also outperforming previous open-source and R1-based video models. These results underscore the effectiveness of our \DATA\ collection and training strategy, reflecting the model’s robust learning capabilities. Crucially, these findings indicate that strengthening an MLLM’s grasp of visual knowledge as a bridge can consistently boost performance across both perceptual grounding and higher-level cognitive reasoning tasks.

\subsection{Ablation Study}
\input{tables/exp_ablation_component}
\paragraph{\textbf{Contribution of Training Strategy.}}
As shown in Table~\ref{tab:exp_ablation_component}, using \textit{See--Think--Answer} SFT alone results in a notable drop in the MS and SI dimensions, likely because merely memorizing visual knowledge does not generalize to diverse and complex social scenarios. Experiments also show that GRPO, with and without $r_v$, tends to repetitive description in the SA, struggling with structured output. Only the two-stage combination of SFT and GRPO achieves consistent gains across all dimensions, with $r_v$ providing an additional 2\% improvement. SFT establishes a foundation for generating structured outputs, while RL enables the model to generalize across diverse visual knowledge scenarios~\cite{chu2025sftmemorizesrlgeneralizes}. 

\begin{table}[t]
    \centering
    \caption{\textbf{Different verifier models comparison.} Qwen2.5-VL-7B performs \textbf{better}.}
    \label{tab:exp_ablation_verifier}
    
    \resizebox{.8\linewidth}{!}{
        \begin{tabular}{ccccccccccc}
        \toprule
        \textbf{Verifier Model} &  \textbf{Overall} &  \textbf{IP} & \textbf{OA} & \textbf{OM} & \textbf{SA} & \textbf{EA}& \textbf{MS} & \textbf{SR} & \textbf{SI}  \\
        \midrule
        Zero-Shot & 63.99 & 48.00 & 53.00 & 82.63 & 35.23 & 75.13 & 77.50 & 82.00 & 65.64 \\
        Qwen2.5-7B & 66.49 & 51.00 & 59.50 & 83.16 & \textbf{39.38} & 79.47 & 78.68 & 80.00 & 66.67 \\
        \rowcolor{mygreen} 
        Qwen2.5-VL-7B & \textbf{67.74} & \textbf{58.00} & \textbf{60.50} & \textbf{83.68} & 38.34 & \textbf{81.05} & \textbf{79.18} & \textbf{80.83} &\textbf{66.92} \\
        \bottomrule
        \end{tabular}
    }
\end{table}

\paragraph{\textbf{Choice of Verifier Model.}} 
We examined the impact of the verifier model on \BENCH\ performance by using both Qwen2.5-VL-7B and its LLM-only counterpart, Qwen2.5-7B~\cite{qwen2.5} to compute the visual knowledge reward. As shown in Table~\ref{tab:exp_ablation_verifier}, using MLLM to verify performs slightly better. Using the same verifier as the basemodel better simulates how MLLM leverages the internal visual knowledge acquired by itself. While a stronger LLM verifier model might yield higher gains, this would diverge from our goal of encouraging reliance more on the models' own visual knowledge rather than on superior language reasoning.
For \textbf{efficiency}, the verifier is externally served via \textit{vLLM} with asynchronous queries, with its latency mostly overlapping other training operations. Empirically, computing $r_v$ adds 1 min (0.3\%) per 1K steps, incurring negligible overhead.

\begin{table}[t]
    \centering
    \caption{\textbf{Sensitivity analysis of $\lambda$.} $\lambda = 0.1$ achieves the \textbf{best} performance.}
    \label{tab:REBUTTAL_exp_ablation_ratio}
    
    \resizebox{.45\linewidth}{!}{
        \begin{tabular}{c|cccccc}
        \toprule
        \textbf{$\lambda$} & \textbf{0.0} & \textbf{0.1} & \textbf{0.3} & \textbf{0.5} & \textbf{0.7} & \textbf{1.0} \\
        \midrule
        \textbf{\BENCH} & 66.0 & \cellcolor{mygreen}\textbf{66.8} & 65.7 & 65.8 & 65.8 & 64.6 \\
        \textbf{MMVU} & 65.9 & \cellcolor{mygreen}67.5 & 66.1 & 66.9 & 65.3 & \textbf{67.8} \\
        \textbf{Video-MME} & 58.0 & \cellcolor{mygreen}58.6 & 58.7 & 58.4 & \textbf{59.1} & 55.5 \\
        \textbf{MVBench} & 63.4 & \cellcolor{mygreen}\textbf{64.8} & 63.9 & 63.1 & 62.6 & 62.9 \\
        \textbf{VSI-Bench} & 35.4 & \cellcolor{mygreen}\textbf{35.9} & 35.7 & 33.6 & \textbf{35.9} & 34.7 \\
        \bottomrule
        \end{tabular}
    }
    \vspace{-3mm}
\end{table}

\paragraph{\textbf{Choice of Visual Knowledge Reward Ratio $\lambda$.}} 
We study the impact of varying $\lambda$, which controls the weight of the visual knowledge reward $r_v$, by training 1K RL steps for rapid exploration. As shown in Table~\ref{tab:REBUTTAL_exp_ablation_ratio}, the best performance \textbf{66.8} is achieved at $\lambda = 0.1$, both omitting the reward and overemphasizing it lead to noticeable drops, indicating that either reliance or disregard of visual knowledge is suboptimal. A moderate incorporation of $r_v$ is crucial for maximizing \MODEL's performance. 
When extended to more benchmarks, different $\lambda$ consistently maintain robust gains, suggesting the effectiveness and generalizability of our reward design.

\subsection{Qualitative Analysis}

\begin{figure}[h]
\centering
\includegraphics[width=\linewidth]{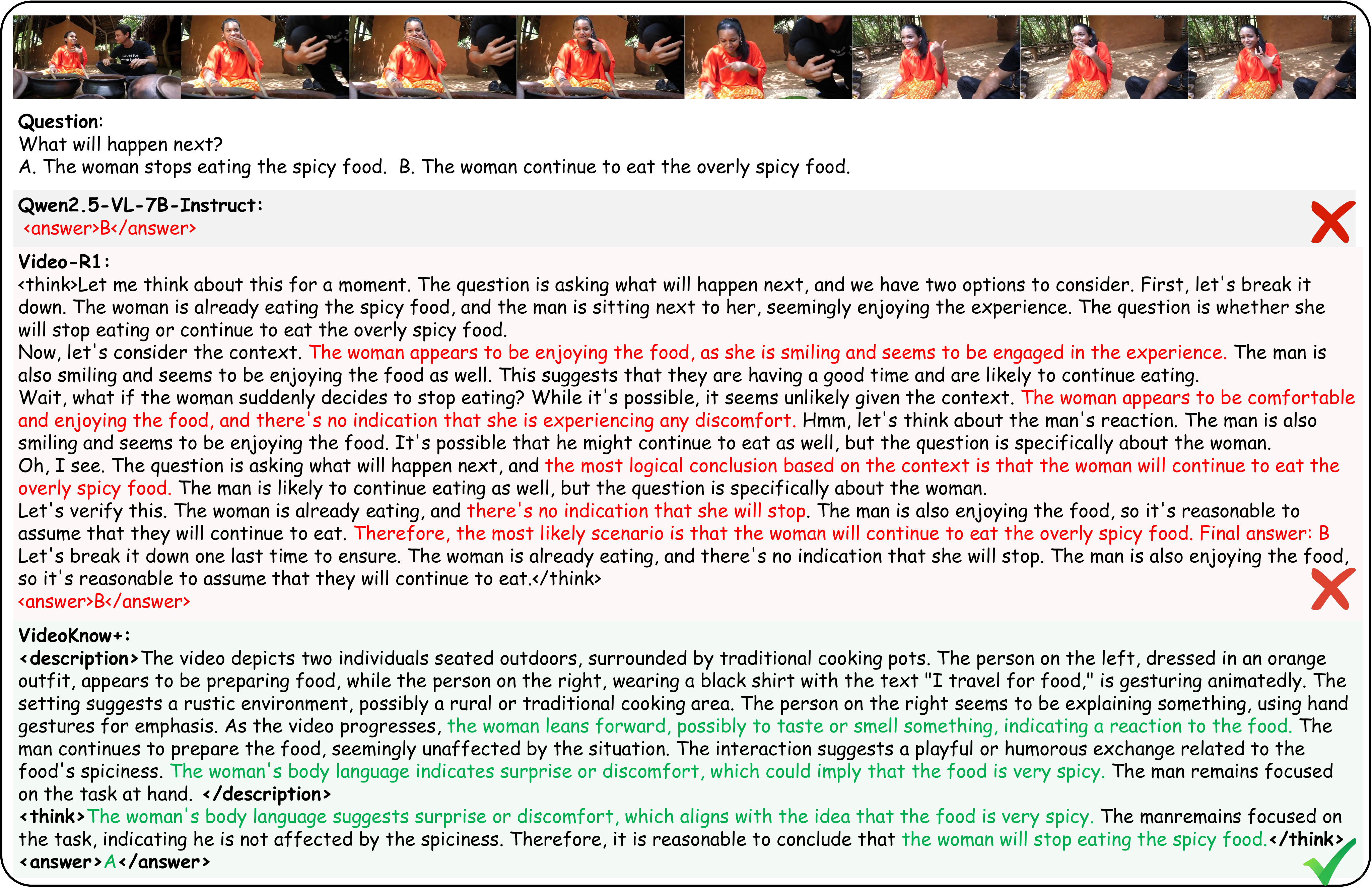}
        \caption{Compared to the Video-R1, \MODEL's decoupled structure produces concise and visually-grounded reasoning, achieving superior stability and accuracy.}  
        \label{fig:quality2}
        \vspace{-5mm}
\end{figure}

Unlike vanilla reasoning which often suffers from shortcuts by bypassing visual processing, our motivation is to first explicitly extract and verify visual knowledge (\textit{See}) before logical deduction (\textit{Think/Answer}). This approach functions as a \textit{think-with-images} variant~\cite{zheng2025deepeyes,su2025thinking} rather than relying on purely text deliberation. 
For instance~(Figure~\ref{fig:quality2}), Video-R1 produces a lengthy reasoning chain with several rounds of self-reflection, yet it still reaches an incorrect conclusion. This discrepancy illustrates the inherent limitations of the traditional \textit{Think–Answer} paradigm, as it often relies too heavily on linguistic reasoning while not sufficiently attention is paid to the extraction of visual knowledge. In contrast, \MODEL\ attains better performance. It generates concise reasoning chains grounded in the visual information obtained directly from the video. This approach leads to outputs that are more coherent, stable, and dependable.
A detailed case study in Appendix G includes more representative examples to provide a more direct understanding of both \BENCH\ and \MODEL.

%% file: tables/exp_ablation_component.tex
\begin{table}[h]
\vspace{-3mm}
\caption{\textbf{Ablation of training strategies on \BENCH.} Combining SFT, GRPO and the visual knowledge reward $r_v$ performs the \textbf{best}.}
\label{tab:exp_ablation_component}
\centering
\resizebox{0.8\columnwidth}{!}{
\begin{tabular}{ccccccccccccc}
\toprule
\textbf{SFT} & \textbf{GRPO} & \textbf{$r_v$} & \textbf{Overall} &  \textbf{IP} & \textbf{OA} & \textbf{OM} & \textbf{SA} & \textbf{EA}& \textbf{MS} & \textbf{SR} & \textbf{SI}  \\
\midrule
\rowcolor{gray!30}
 &  &  & 63.99 & 48.00 & 53.00 & 82.63 & 35.23 & 79.47 & 75.13 & 77.50 & 65.64 \\
\checkmark & & & 63.87 & 57.00 & 54.50 & 80.00 & \textbf{43.52} & 78.95 & 69.54 & \textbf{81.67} & 58.72 \\
 & \checkmark & &  64.29 & 48.50 & 57.50 & 82.11 & 31.61 & 77.89 & 78.17 & 83.33 & 63.85\\
 & \checkmark & \checkmark & 63.63 & 53.50 & 56.50 & 82.11 & 22.28 & \textbf{81.58} & 77.16 & 80.83 & 63.08 \\
 \checkmark & \checkmark &  & 65.71 & 52.50 & 59.50 & 83.16 & 42.49 & 78.42 & 78.68 & 79.17 & 61.79 \\
\rowcolor{mygreen} \checkmark & \checkmark & \checkmark & \textbf{67.74} & \textbf{58.00} & \textbf{60.50} & \textbf{83.68} & 38.34 & 81.05 & \textbf{79.19} & 80.83 & \textbf{66.92} \\

\bottomrule
\end{tabular}
}
\vspace{-3mm}
\end{table}

%% file: sec/6_Conclusion.tex
\section{Conclusion}
In this paper, we highlight the critical role of visual knowledge in the development of MLLMs, which encompasses concepts rooted in human cognitive psychology and serves to bridge perception and reasoning. To measure this semantics quantitatively, we introduce \textbf{\BENCH}, a comprehensive video benchmark designed to evaluate MLLMs’ understanding of visual knowledge across both world-centric and human-centric scenarios. Furthermore, we propose \textbf{\MODEL}, an initial attempt to explicitly integrate visual knowledge into MLLMs with \textbf{\DATA}, achieving notable performance on \BENCH\ as well as other video benchmarks.  

Future work could explore harder QA formats beyond multiple choice and higher-quality datasets tailored to visual knowledge.

%% file: sec/Supplementary_Material.tex
\clearpage
\appendix
\renewcommand{\thesection}{\Alph{section}}
\setcounter{section}{0}

\titlerunning{VKnowU}

\author{Tianxiang Jiang\inst{1,2}\orcidlink{0009-0001-3957-9748} \and
Sheng Xia\inst{3,4}\orcidlink{0009-0005-8889-3927} \and
Yicheng Xu\inst{2}\orcidlink{0000-0003-2975-1206} \and
Linquan Wu\inst{5}\orcidlink{0009-0000-6594-625X} \and \\
Xiangyu Zeng\inst{3,2}\orcidlink{0000-0001-6956-5040} \and
Limin Wang\inst{3,2}\orcidlink{0000-0002-3674-7718} \and
Yu Qiao\inst{2}\orcidlink{0000-0002-1889-2567} \and
Yi Wang\inst{2}\thanks{Corresponding author.}\orcidlink{0000-0001-9134-1203}}

\authorrunning{T.~Jiang et al.}

\institute{
$ ^{1}$University of Science and Technology of China \quad 
$ ^{2}$Shanghai AI Laboratory \\
$ ^{3}$Nanjing University 
$ ^{4}$Shanghai Innovation Institute 
$ ^{5}$City University of Hong Kong
\\
\email{
\url{https://github.com/OpenGVLab/VKnowU}
}
}

\title{Supplementary Material of\\
VKnowU: Evaluating Visual Knowledge Understanding in Multimodal LLMs}
\maketitle
  \renewcommand{\thefigure}{A\arabic{figure}}
  \renewcommand{\thetable}{A\arabic{table}}
  
\startcontents[sections]
\begingroup
    \hypersetup{linkcolor=black}
    \renewcommand{\baselinestretch}{1.5}\selectfont
    \printcontents[sections]{l}{1}{\setcounter{tocdepth}{2}}
\endgroup
\newpage

\section{Theoretical Analysis of \MODEL}
\label{sec:appendix_theoretical_analysis}

The reasoning process of multimodal large language models (MLLMs) can be formulated as conditional probability inference grounded in visual evidence. Given a visual input $\mathbf{V}$ (e.g., images or videos), a user query $\mathbf{T}_q$, and the corresponding ground-truth answer $\mathbf{T}_a$, the reasoning objective of an MLLM can be expressed as:

\begin{equation}
    \underbrace{P(\mathbf{T}_a | \mathbf{V}, \mathbf{T}_q)}_{\text{VQA accuracy}}
    \propto
    \underbrace{P(\mathbf{V} | \mathbf{T}_a, \mathbf{T}_q)}_{\text{visual evidence}}
    \cdot
    \underbrace{P(\mathbf{T}_a | \mathbf{T}_q)}_{\text{language prior}} .
\end{equation}

Here, $P(\mathbf{V} \mid \mathbf{T}_a, \mathbf{T}_q)$ represents how well the visual input supports the candidate answer given the query. When the visual representation contains rich and precise cues relevant to $\mathbf{T}_a$, this term becomes larger, indicating stronger grounding for the prediction. In contrast, $P(\mathbf{T}_a \mid \mathbf{T}_q)$ reflects the language prior acquired during large-scale pretraining. When the visual evidence is ambiguous, noisy, or insufficient, MLLMs may rely heavily on this prior and produce plausible but visually unsupported predictions.

Accordingly, the final prediction can be written as:

\begin{equation}
    \hat{\mathbf{T}}_a
    =
    \arg \max_{\mathbf{T}_a}
    P(\mathbf{V} | \mathbf{T}_a, \mathbf{T}_q)
    \cdot
    P(\mathbf{T}_a | \mathbf{T}_q).
\end{equation}

This formulation motivates \MODEL: by explicitly strengthening the visual grounding term through the \textit{See--Think--Answer} format and the visual knowledge reward, the model is encouraged to base its answers on visual evidence rather than shortcutting through language priors. This helps reduce hallucination and improves visual knowledge understanding.

\section{Principle of Group Relative Policy Optimization}
\label{sec:appendix_grpo}

Group Relative Policy Optimization (GRPO)~\cite{shao2024deepseekmath} has shown strong effectiveness across both textual and multimodal tasks~\cite{huang2025vision,feng2025video}. Compared with PPO-style methods, GRPO estimates the baseline directly from group-level reward statistics, avoiding an additional value network and reducing training overhead.

For each query $q$, the old policy $\pi_{\theta_{\mathrm{old}}}$ samples a group of $G$ outputs $\{o_1,\dots,o_G\}$. These outputs are evaluated by reward functions, including the format reward, the answer reward, and the visual knowledge reward defined in the main paper. The advantage of the $i$-th output is computed by group normalization:

\begin{equation}
    A_i =
    \frac{R_i - \mathrm{mean}(\{R_j\}_{j=1}^{G})}
    {\mathrm{std}(\{R_j\}_{j=1}^{G})},
\end{equation}
where $R_i$ is the total reward of $o_i$. The GRPO objective is:

\begin{equation}
\begin{aligned}
\mathcal{J}_{\mathrm{GRPO}}(\theta)
&=
\mathbb{E}_{q,\{o_i\}\sim \pi_{\theta_{\mathrm{old}}}}
\Bigg[
\frac{1}{G}\sum_{i=1}^{G}
\min\Big(
r_i(\theta) A_i,\,
\mathrm{clip}(r_i(\theta),1-\epsilon,1+\epsilon) A_i
\Big) \\
&\qquad
- \beta\,\mathbb{D}_{\mathrm{KL}}
\big(\pi_\theta(\cdot|q)\,\|\,\pi_{\mathrm{ref}}(\cdot|q)\big)
\Bigg],
\end{aligned}
\end{equation}
where
\begin{equation}
    r_i(\theta)=
    \frac{\pi_\theta(o_i\mid q)}
    {\pi_{\theta_{\mathrm{old}}}(o_i\mid q)} .
\end{equation}
Here, $\epsilon$ controls the clipping range and $\beta$ controls the KL regularization strength.

\section{Details of \BENCH}
\label{sec:appendix_vknowu_details}

\subsection{Data Provenance and Filtering Statistics}
\label{sec:appendix_provenance}

Table~\ref{tab:appendix_provenance} summarizes the annotation sources, video sources, candidate-pool size, retained sample count, filtering rate, and rewriting rate for each task in \BENCH. The collected datasets cover diverse physical and social scenarios, providing reliable initial annotations for constructing the final benchmark. Overall, \BENCH\ retains only 1,680 out of 42,670 candidates (3.9\%), showing that the final benchmark is a heavily filtered reformulation benchmark rather than a direct aggregation of existing annotations. All human-centric items are rewritten to reduce shortcut solutions and improve distractor quality.

\begin{table}[h]
\caption{Data provenance and filter/rewrite rate of \BENCH.}
\label{tab:appendix_provenance}
\centering
\resizebox{\linewidth}{!}{
\begin{tabular}{c|c|c|c|c|c|c}
\toprule
\textbf{Task} & \textbf{Annotation Source} & \textbf{Video Source} & \textbf{Pool} & \textbf{Kept} & \textbf{Filter} & \textbf{Rewrite} \\
\midrule
\textbf{Intuitive Physics} & IntPhys2~\cite{bordes2025intphys} & Unreal Engine~\cite{unrealengine} & 1{,}416 & 200 & 14.1\% & -- \\
\textbf{Object Affordance} & PACS~\cite{yu2022pacs} & YouTube & 13{,}400 & 200 & 1.5\% & -- \\
\textbf{Object Material} & PACS~\cite{yu2022pacs} & YouTube & 4{,}349 & 190 & 4.4\% & -- \\
\textbf{Spatial Awareness} & VSI-Bench~\cite{yang2025thinkingspacemultimodallarge} & ARKitScenes~\cite{baruch2021arkitscenes} & 5{,}130 & 193 & 3.8\% & -- \\
\textbf{Event Anticipation} & VLEP~\cite{lei2020more} & YouTube vlog & 4{,}192 & 190 & 4.5\% & \textbf{100\%} \\
\textbf{Mental State} & Social-IQ 2.0~\cite{siq2} & YouTube & 6{,}020 & 197 & 3.3\% & \textbf{100\%} \\
\textbf{Social Relation} & Social-IQ 2.0~\cite{siq2} & YouTube & 6{,}020 & 120 & 2.0\% & \textbf{100\%} \\
\textbf{Subjective Intention} & RexTime~\cite{chen2024rextime} & QVHighlights~\cite{lei2021detecting}, ActivityNet~\cite{caba2015activitynet} & 2{,}143 & 390 & 18.2\% & \textbf{100\%} \\
\midrule
\textbf{Total} & \textbf{\BENCH} & \textbf{--} & \textbf{42{,}670} & \textbf{1{,}680} & \textbf{3.9\%} & \textbf{53.4\%} \\
\bottomrule
\end{tabular}
}
\end{table}

\subsection{Correlation with Existing Benchmarks}
\label{sec:appendix_benchmark_correlation}

\begin{figure}[t!]
    \centering
    \includegraphics[width=1\linewidth]{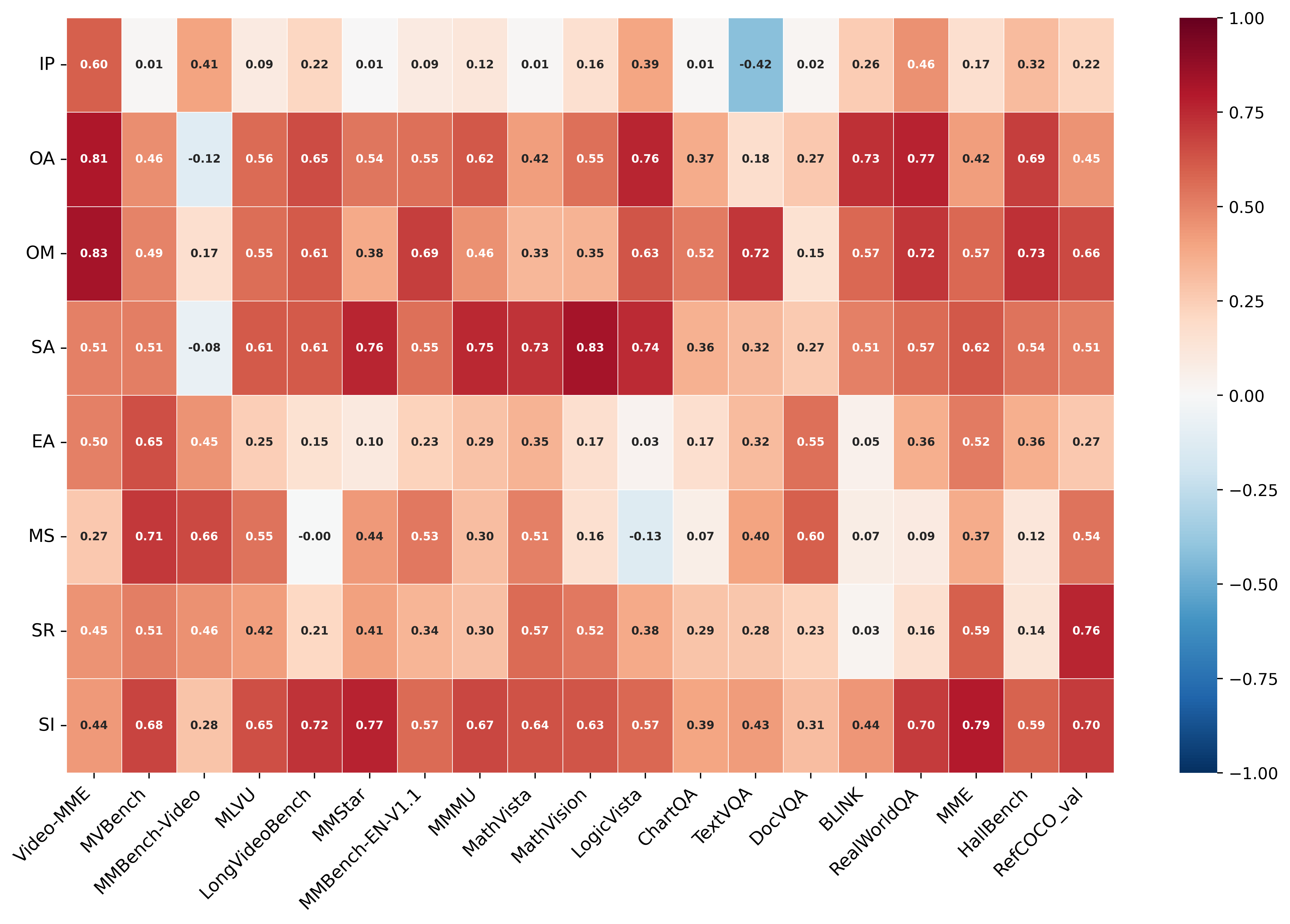}
    \caption{Pearson correlations between the visual knowledge dimensions in \BENCH\ and existing MLLM benchmarks.}
    \label{fig:appendix_correlation_between_others}
\end{figure}

We further compute Pearson correlations between the eight \BENCH\ sub-tasks and a diverse set of established MLLM benchmarks to examine how \BENCH\ relates to existing evaluation suites. As shown in Fig.~\ref{fig:appendix_correlation_between_others}, the correlation structure reveals two consistent patterns.

First, tasks such as Spatial Awareness (SA) and Object Material (OM) show moderate-to-strong alignment with perception-oriented benchmarks, including MMBench~\cite{liu2024mmbench}, MMStar~\cite{chen2024we}, and MME~\cite{fu2025mmecomprehensiveevaluationbenchmark}. These benchmarks contain object-centric and scene-centric signals, which partially overlap with the perceptual competencies required by SA and OM. In contrast, Intuitive Physics (IP) and Event Anticipation (EA) exhibit consistently weak correlations with most existing benchmarks, including video-based ones. This suggests that physical and temporal visual knowledge remains underrepresented in current evaluation pipelines.

Second, on the human-centric side, Social Relation (SR) and Subjective Intention (SI) show moderate-to-strong correlations with grounding-intensive benchmarks such as HallusionBench~\cite{guan2024hallusionbench} and RefCOCO~\cite{yu2016modeling}. Although these datasets do not explicitly target social or relational reasoning, they require accurate grounding of humans, body poses, actions, and spatial configurations, which are also important cues for SR and SI. Mental State (MS) exhibits weaker correlations, suggesting that existing datasets capture only part of the capability required for mental-state inference.

Overall, \BENCH\ is not entirely orthogonal to existing benchmarks. Instead, it exposes a clear imbalance: current evaluation suites emphasize perception and grounding more than world-centric intuitive knowledge. This highlights the diagnostic value of \BENCH\ for identifying missing dimensions in current MLLM evaluation.

\section{Details of \DATA}
\label{sec:appendix_vknowqa_details}

\subsection{\DATA-30K Data Sources}

\DATA-30K is constructed from open-source video QA and reasoning datasets. We organize the data into three categories: general video understanding, world-centric visual knowledge, and human-centric visual knowledge. The source composition is summarized in Table~\ref{tab:traindata_source}.

\input{tables/traindata_source}

\subsection{\DATA-CS-12K Generation Pipeline}
\label{sec:appendix_cs12k_pipeline}

To obtain cold-start data for the \textit{See--Think--Answer} format, we first prompt Qwen2.5-VL-7B-Instruct to generate structured responses, following the template in Fig.~\ref{fig:prompt_see_think_answer}. We retain only the instances where the model produces both the correct answer and the desired output format.

We then apply an additional filtering step using the verifier prompt in Fig.~\ref{fig:prompt_reward}. Specifically, the generated visual description is treated as a text-only proxy for the visual input. We keep an instance only if the model can correctly answer the question from this description, which ensures that the description contains sufficient explicit visual knowledge rather than relying on hidden priors. After these two refinement steps, we obtain approximately 12K high-quality QA pairs, denoted as \DATA-CS-12K.

\subsection{Data Statistics}
\label{sec:appendix_data_statistics}

\begin{figure}[h]
    \centering
    \begin{subfigure}[b]{0.32\linewidth}
        \centering
        \includegraphics[width=\linewidth]{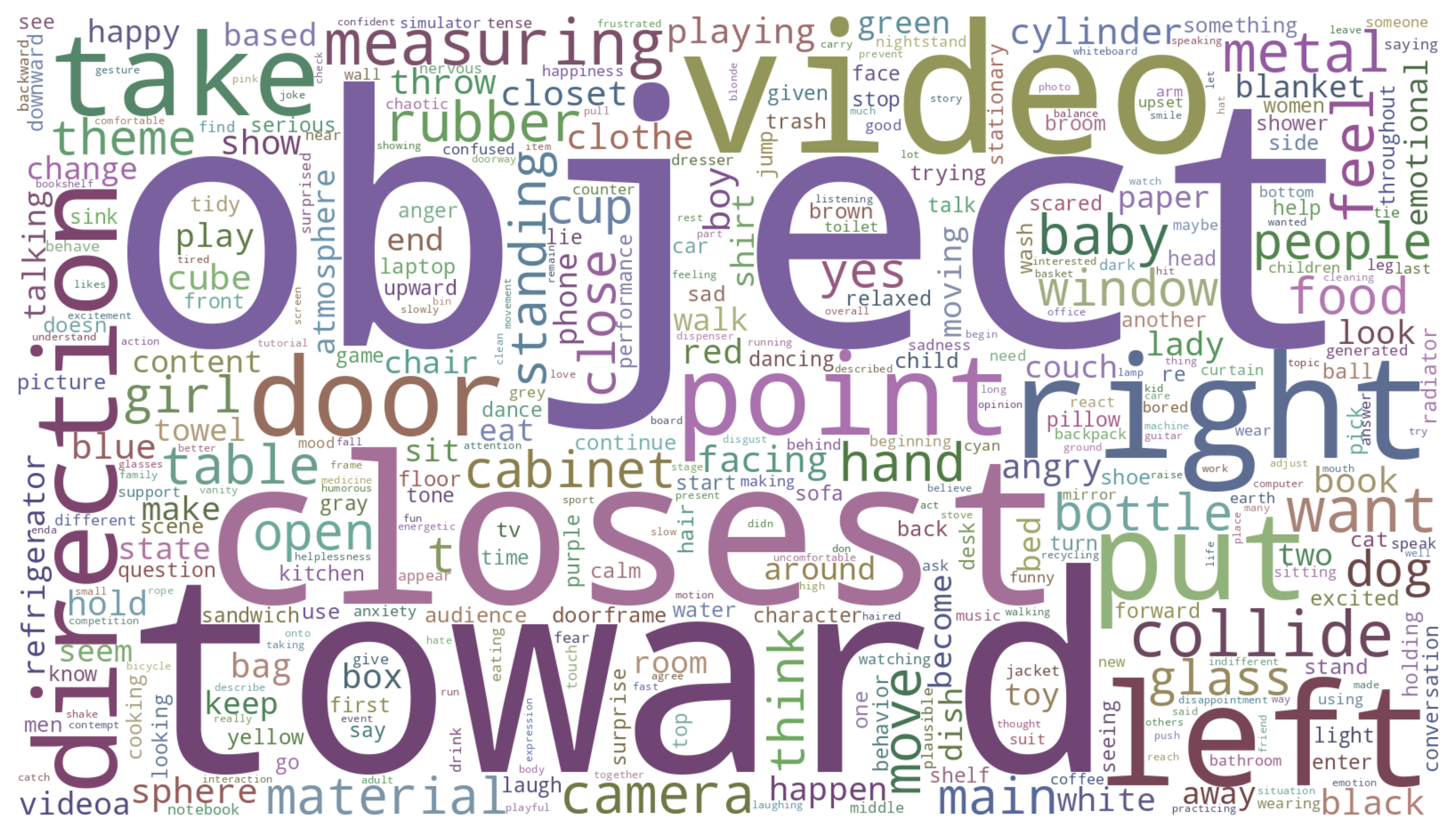}
        \caption{\DATA-30K}
        \label{fig:appendix_30k_question_wordcloud}
    \end{subfigure}
    \hfill
    \begin{subfigure}[b]{0.32\linewidth}
        \centering
        \includegraphics[width=\linewidth]{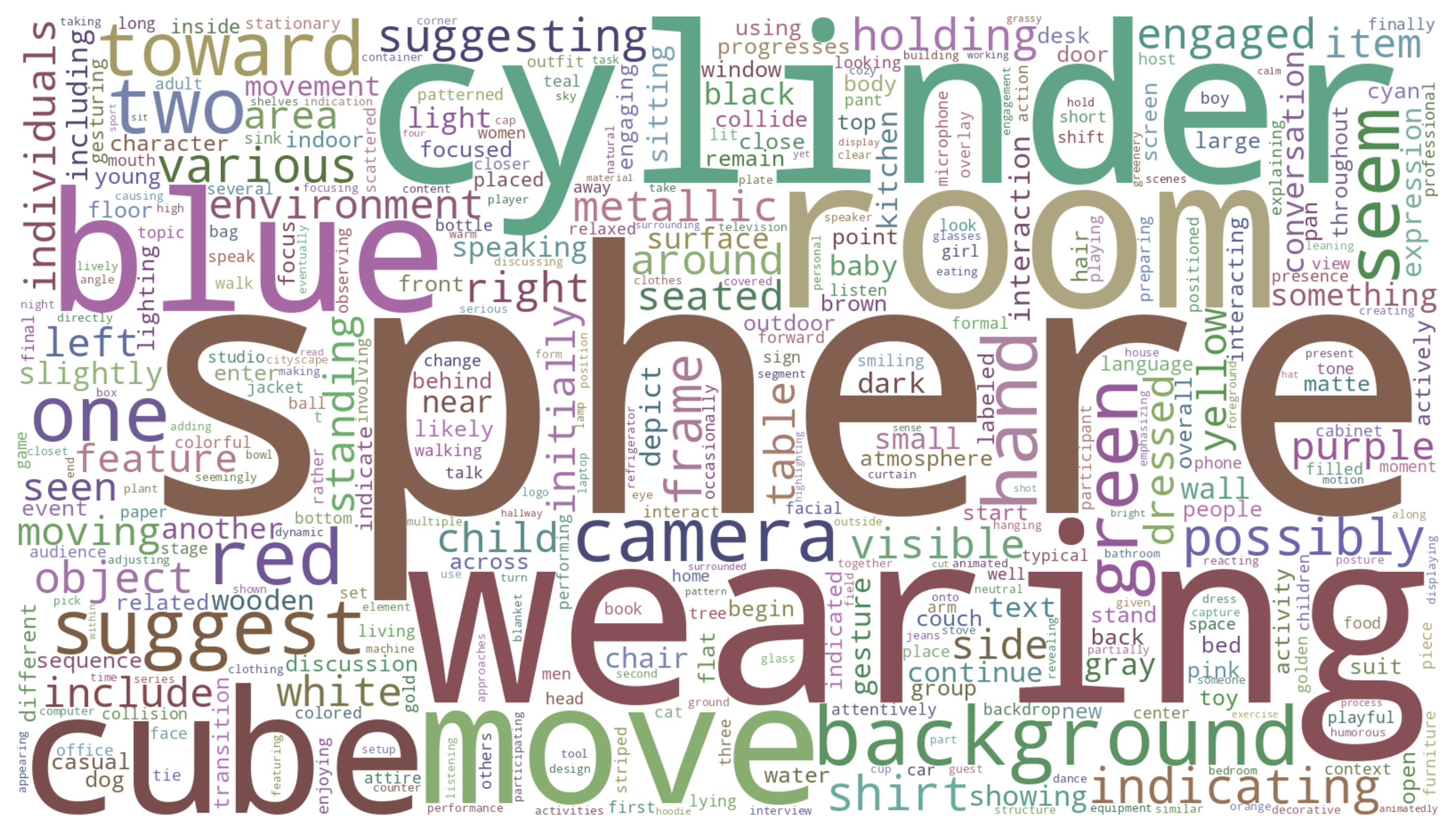}
        \caption{CS-12K \textit{description}}
        \label{fig:appendix_12k_description_wordcloud}
    \end{subfigure}
    \hfill
    \begin{subfigure}[b]{0.32\linewidth}
        \centering
        \includegraphics[width=\linewidth]{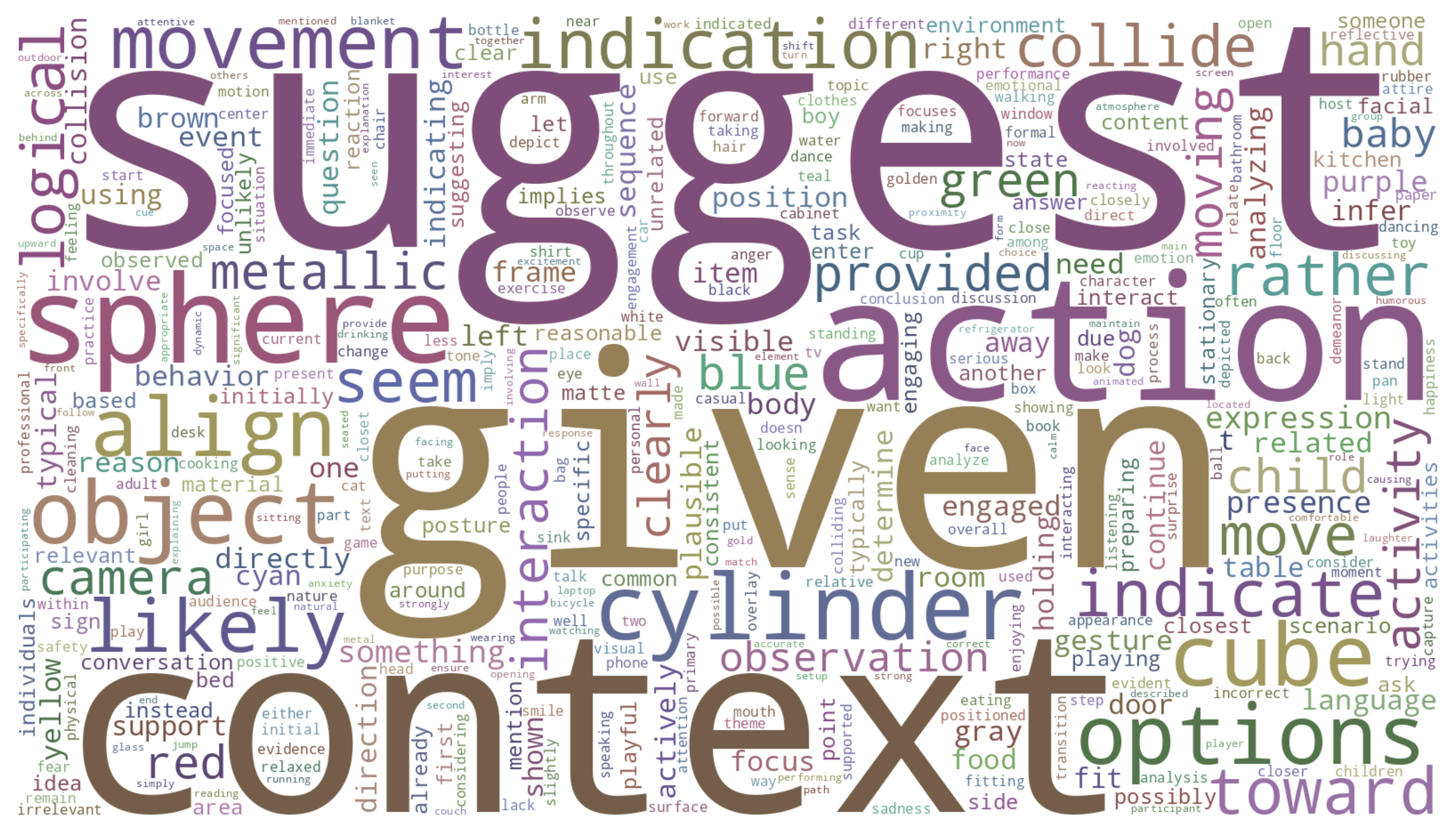}
        \caption{CS-12K \textit{think}}
        \label{fig:appendix_12k_think_wordcloud}
    \end{subfigure}
    \caption{Word cloud visualization of \DATA.}
    \label{fig:appendix_wordcloud_visualization}
\end{figure}

Figure~\ref{fig:appendix_wordcloud_visualization} visualizes the lexical distribution of questions in \DATA-30K and the generated \textit{description} and \textit{think} fields in \DATA-CS-12K. The word clouds indicate that the data cover both object-level visual cues and higher-level reasoning concepts, consistent with the world-centric and human-centric design of \BENCH.

\section{Implementation Details}
\label{sec:appendix_implementation}

\subsection{Training}
\label{sec:appendix_training_details}

\paragraph{\textbf{Baselines.}}
Video-R1~\cite{feng2025video} and VideoRFT~\cite{wang2025videorft} are used as reasoning-oriented video MLLM baselines. Video-R1 explores an R1-style video reasoning pipeline with SFT cold start and T-GRPO training, while VideoRFT follows a related training style with higher-quality CoT data and a semantic-consistency reward.

\begin{figure}[htbp]
    \centering
    \includegraphics[width=.73\linewidth]{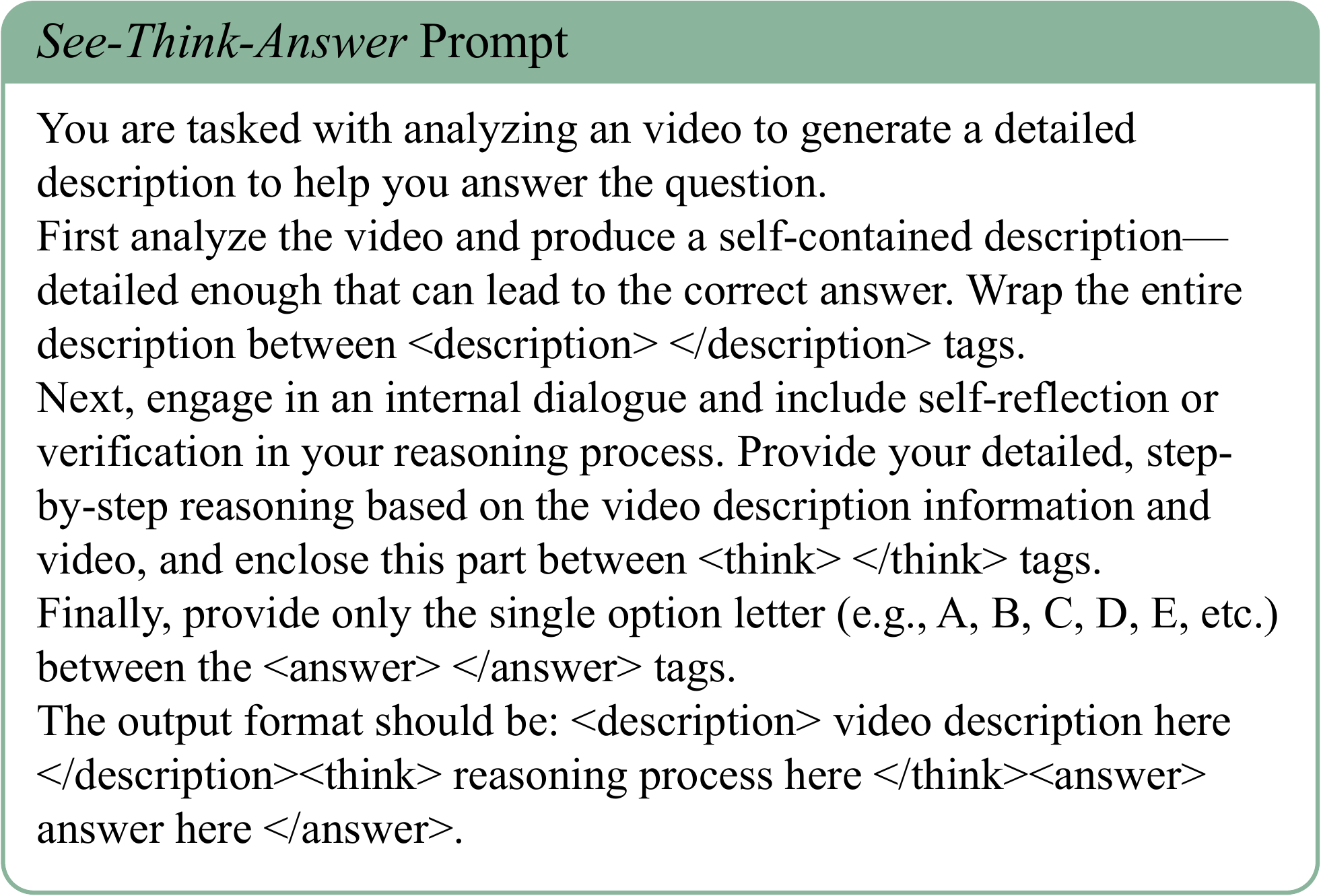}
    \caption{Prompt for MLLMs to generate the \textit{See--Think--Answer} output format.}
    \label{fig:prompt_see_think_answer}
\end{figure}

\begin{figure}[htbp]
    \centering
    \includegraphics[width=.9\linewidth]{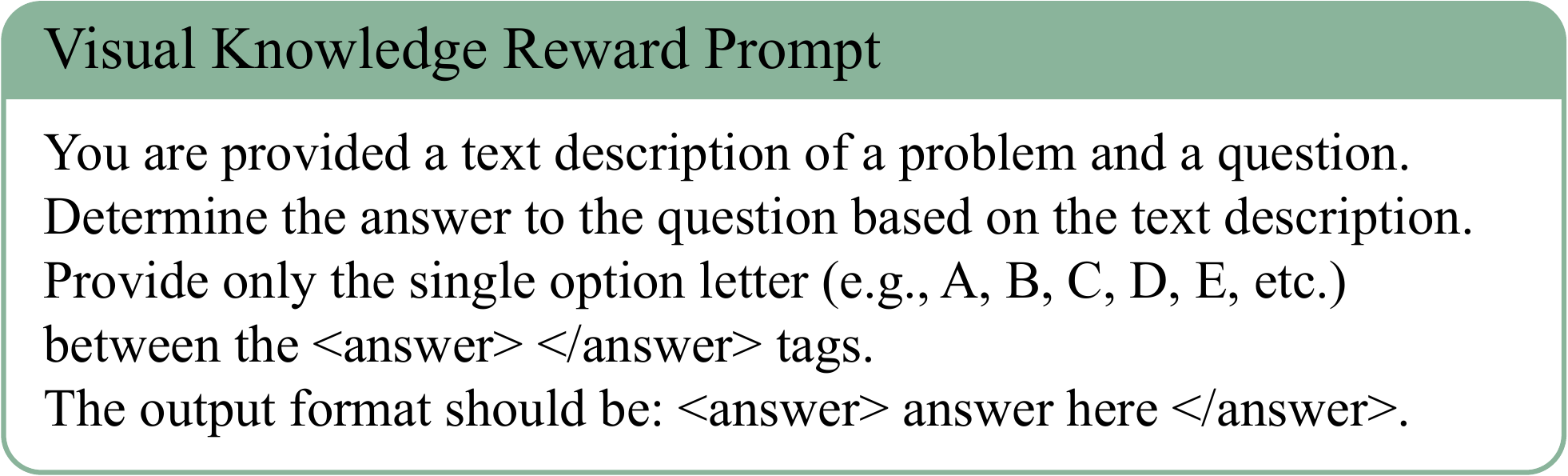}
    \caption{Prompt for the verifier model to calculate the visual knowledge reward.}
    \label{fig:prompt_reward}
\end{figure}

\paragraph{\textbf{Prompts.}}
We use the prompt in Fig.~\ref{fig:prompt_see_think_answer} to guide the model to generate \textit{See--Think--Answer} responses during both SFT and GRPO. The verifier prompt in Fig.~\ref{fig:prompt_reward} is used to compute the binary visual knowledge reward $r_v$.

\paragraph{\textbf{Verifier Serving.}}
The visual knowledge reward is computed by a frozen Qwen2.5-VL-7B verifier. During RL training, the verifier is externally served through \textit{vLLM} with asynchronous queries, so most verifier latency overlaps with other training operations. Empirically, computing $r_v$ adds about 1 minute per 1K training steps, corresponding to roughly 0.3\% extra overhead.

\paragraph{\textbf{Hyperparameters.}}
Our training implementation is based on the TRL framework~\cite{vonwerra2022trl}. Table~\ref{tab:appendix_training_hyperparameters} reports the hyperparameters used for the main experiments. To balance training efficiency and compute cost, we follow prior work~\cite{huang2025vision,wang2025videorft} and limit the maximum number of video frames to 16 during training. Each frame is processed at a resolution of up to $128 \times 28 \times 28$ pixels. All experiments are reproducible using 8 NVIDIA A800 GPUs.

\begin{table}[h!]
\caption{Training hyperparameters.}
\label{tab:appendix_training_hyperparameters}
\centering
\resizebox{.8\columnwidth}{!}{%
\begin{tabular}{lcc}
\toprule
\textbf{Parameter} & \textbf{SFT} & \textbf{GRPO} \\
\midrule
train type & full & full \\
use vLLM & false & true \\
vLLM GPU memory utilization & - & 0.7 \\
attention implementation & flash\_attn2 & flash\_attn2 \\
DeepSpeed config & zero2 & zero3 \\
torch dtype & bfloat16 & bfloat16 \\
num train epochs & 1 & 1 \\
per-device train batch size & 1 & 1 \\
gradient accumulation steps & 2 & 1 \\
num generations & - & 8 \\
KL coefficient $\beta$ & - & 0.04 \\
visual knowledge reward ratio $\lambda$ & - & 0.1 \\
learning rate & 1e-6 & 1e-6 \\
max prompt length & 16384 & 16384 \\
max completion length & 1024 & 1024 \\
max grad norm & 5.0 & 5.0 \\
min frames & 4 & 4 \\
max frames & 16 & 16 \\
video pixels & $128 \times 28 \times 28$ & $128 \times 28 \times 28$ \\
\bottomrule
\end{tabular}
}
\end{table}

\subsection{Evaluation}
\label{sec:appendix_evaluation_details}

The models evaluated in our study vary significantly in architecture and scale. All experiments are conducted on NVIDIA A800 GPUs with 80 GB memory. To ensure reproducibility, we follow the official implementations and configurations released by the model developers.

\begin{figure}[htbp]
    \centering
    \includegraphics[width=.73\linewidth]{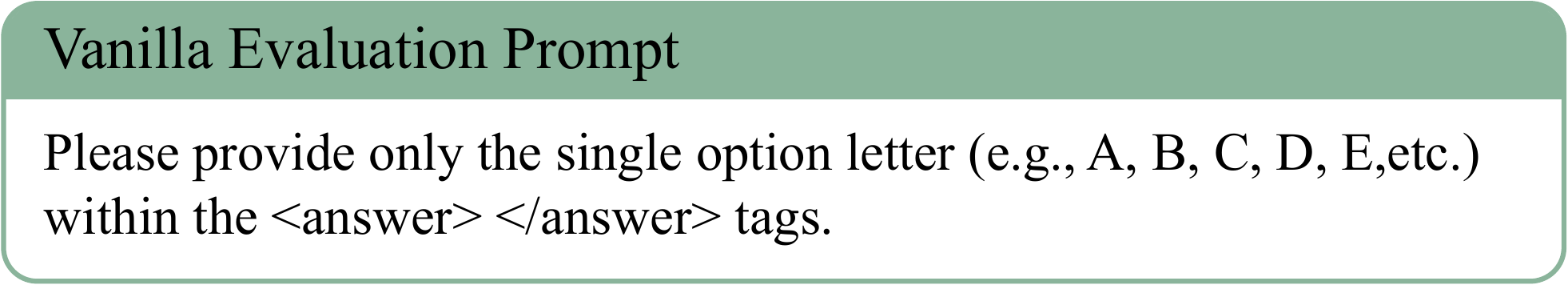}
    \caption{Prompt for \BENCH\ evaluation for vanilla models.}
    \label{fig:prompt_vanilla}
\end{figure}

\paragraph{\textbf{Prompts.}}
For models that do not require explicit thinking, we use the prompt in Fig.~\ref{fig:prompt_vanilla} to guide concise responses. For \MODEL, we use the same \textit{See--Think--Answer} prompt as in training. For other thinking models, we follow the official prompts specified by their developers.

\paragraph{\textbf{Hyperparameters.}}
For QwenVL-based models, including Qwen2.5-VL, MiMo-VL, and \MODEL, we evaluate with video frame counts ranging from 4 to 32 and fix the video resolution to $256 \times 28 \times 28$ pixels. For other open-source and proprietary models, we fix the maximum number of input video frames to 32 for comparability. Except for proprietary models, which are evaluated with temperature 1.0, all open-source models are evaluated with temperature 0.1 and top-$p$ 0.001.

\subsection{Additional Training Variants}
\label{sec:appendix_additional_training_variants}

The main paper reports the core comparison, training-strategy ablation, verifier comparison, and $\lambda$ sensitivity in Tables 4, 5, 6, and 7. Here, we include additional variants omitted from the main text to avoid repetition.

\begin{table}[h]
\caption{Additional training variants omitted from the main table. Results should be read together with the Zero-Shot and \MODEL\ rows in Table 4.}
\label{tab:appendix_additional_training_variants}
\centering
\resizebox{.9\linewidth}{!}{
\begin{tabular}{lcccccc}
\toprule
\textbf{Model Variant} & \textbf{\BENCH} & \textbf{MVBench} & \textbf{Video-MME} & \textbf{MMVU} & \textbf{VSI-Bench} & \textbf{Avg.} \\
\midrule
+ \textit{See--Think--Answer} prompt & 64.7 & 60.6 & 54.4 & 62.9 & 32.2 & 55.0 \\
+ Answer SFT & 65.3 & 63.0 & 60.4 & 65.6 & 34.2 & 57.7 \\
\quad + GRPO only with $r_a$ & 64.9 & 63.7 & 60.7 & 65.7 & 35.1 & 58.0 \\
\MODEL\ (better captions) & \textbf{68.0} & \textbf{65.8} & 60.3 & \textbf{68.0} & \textbf{36.6} & \textbf{59.8} \\
\bottomrule
\end{tabular}
}
\end{table}

The \textit{See--Think--Answer} prompt alone brings a modest gain over zero-shot inference, suggesting that structured visual grounding is helpful even without training. However, direct answer SFT and GRPO with only the answer reward $r_a$ remain inferior to \MODEL, indicating that answer correctness alone is insufficient for learning robust visual knowledge. As a data-quality diagnostic, we also regenerate captions with Qwen3-VL-235B-A22B and keep the same training recipe. The resulting \MODEL\ variant obtains consistent gains, suggesting that stronger visual descriptions further improve the effectiveness of visual-knowledge-oriented training.

\begin{figure*}[htbp]
    \centering
    \includegraphics[width=0.9\linewidth]{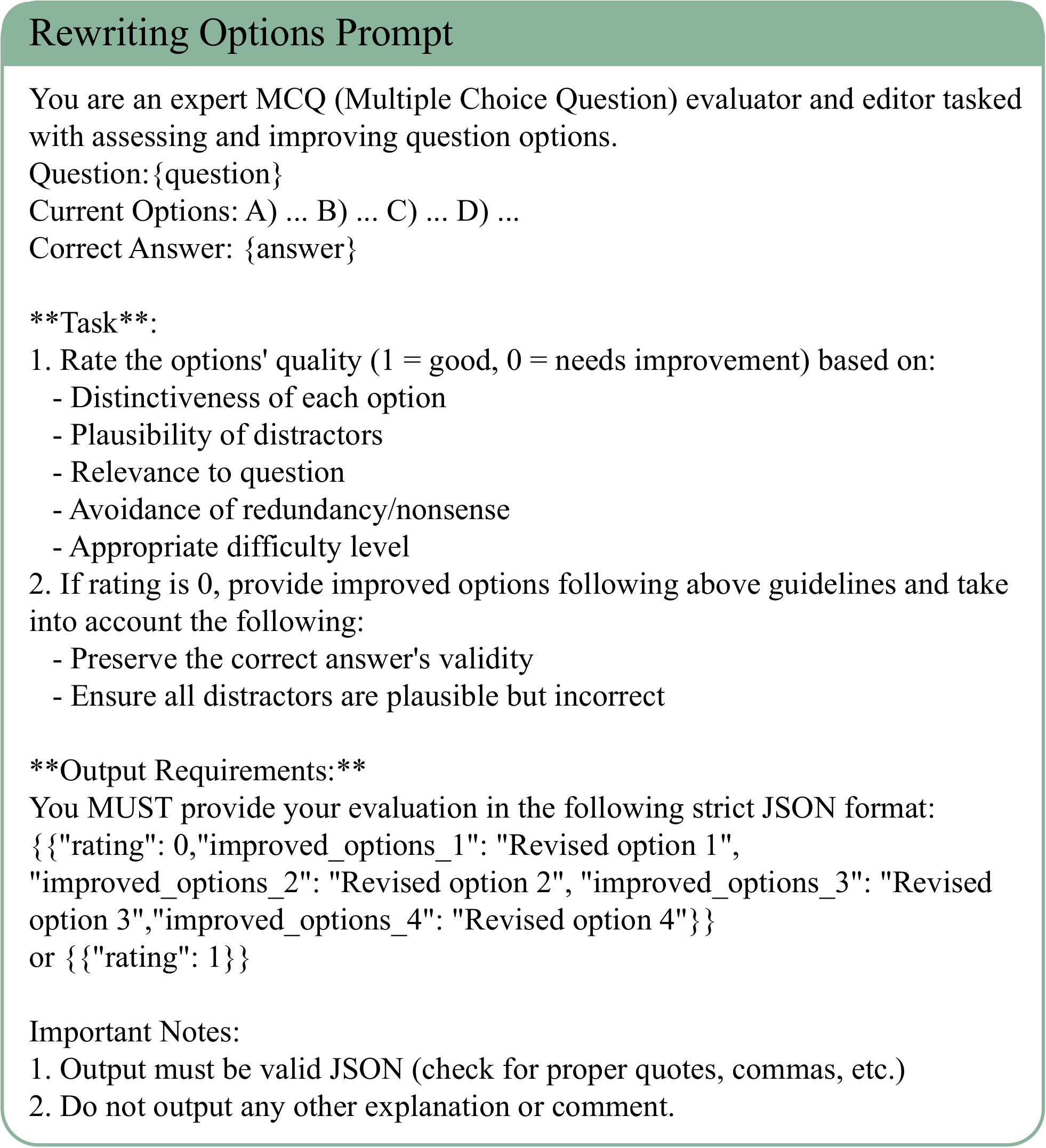}
    \caption{Prompt for DeepSeek-R1~\cite{guo2025deepseek} to enhance distractor options.}
    \label{fig:prompt_rewrite}
\end{figure*}

\clearpage
\section{Scope of the Eight Visual Knowledge Tasks}
\label{sec:appendix_task_scope}

\paragraph{\textbf{Intuitive Physics.}}
Intuitive Physics refers to the ability to judge the physical plausibility of dynamic events by applying common-sense principles such as object permanence, solidity, and continuity~\cite{spelke1992origins,baillargeon1986representing}. This goes beyond recognizing objects: it requires predicting how they should behave under physical laws. For example, if a ball rolls behind a screen, humans expect it to continue its trajectory and emerge from the other side, rather than vanish or pass through a solid obstacle. Such knowledge has been studied in violation-of-expectation experiments with infants~\cite{baillargeon1986representing}.

\paragraph{\textbf{Object Affordance.}}
Object Affordance is the ability to infer an object's potential functional uses from its perceptual and structural properties~\cite{gibson1979ecological}. For example, a flat knee-high surface affords sitting, a small graspable object affords lifting, and a sharp edge affords cutting. This is distinct from object identification: one can perceive an object's ``sittability'' or ``cutting ability'' even without relying solely on its category name. The design of everyday objects often depends on making such affordances visually clear~\cite{norman1988psychology}.

\paragraph{\textbf{Object Material.}}
Object Material concerns inferring material composition from visual cues such as texture, gloss, transparency, and deformation~\cite{fleming2017material,adelson2001seeing}. This ability supports predictions about physical properties such as weight, fragility, and flexibility without direct touch. For instance, a transparent drinking glass is inferred to be rigid and fragile, whereas a paper cup is inferred to be light, flexible, and opaque. Human material perception combines low-level visual cues~\cite{zhao2026luve,zhao2026zerodetailprogressivespectral} with high-level conceptual knowledge~\cite{schmidt2025core}.

\paragraph{\textbf{Spatial Awareness.}}
Spatial Awareness is the ability to understand relative positions, orientations, and relationships within an environment~\cite{o1978hippocampus}. It requires building a mental representation of the scene rather than only identifying objects. For example, a model with spatial awareness should distinguish whether a cat is on or under a mat, or whether a lamp is to the left or right of a sofa.

\paragraph{\textbf{Event Anticipation.}}
Event Anticipation is the ability to infer likely future events from current visual evidence and learned scripts. For example, when seeing a person in a restaurant reading a menu, humans anticipate that the person may order food rather than suddenly leave. This ability reflects predictive processing, where the brain continuously forms and updates hypotheses about future sensory input~\cite{friston2010free,clark2013whatever}.

\paragraph{\textbf{Mental State.}}
Mental State inference, often related to Theory of Mind, is the ability to infer others' beliefs, desires, intentions, and emotions~\cite{premack1978does,baron1985does}. It goes beyond recognizing visible expressions. For example, recognizing a smile as happiness is only one component; inferring that someone is happy because they believe they have won requires a deeper mental-state judgment~\cite{ekman1992argument}. Classic false-belief tests further demonstrate the distinction between observed reality and another agent's internal belief~\cite{wimmer1983beliefs}.

\paragraph{\textbf{Social Relation.}}
Social Relation inference is the ability to identify interpersonal relationships and social dynamics from non-verbal cues such as physical proximity, gaze, posture, and interaction patterns~\cite{hall1966hidden,argyle1965eye}. For instance, two people standing close together with frequent eye contact and open posture may suggest familiarity, whereas large distance, gaze avoidance, and closed posture may indicate a more formal or distant relationship.

\paragraph{\textbf{Subjective Intention.}}
Subjective Intention involves reconstructing the goal or motivation behind an observed action~\cite{dennett1987intentional}. It asks not only what a person is doing, but why they are doing it. For example, if a person repeatedly tries to place a book on a high shelf, the intended goal is to shelve the book even if the action is unsuccessful. This differs from moral evaluation, which judges the rightness of an intention; here, the goal is simply to infer the intended action or motivation~\cite{piaget1965moral,brentano1995psychology}.

\section{Case Study}
\label{sec:appendix_case_study}

We present representative examples to provide a more direct understanding of both \BENCH\ and \MODEL.

\subsection{Strengths}
\label{sec:appendix_case_strengths}

\begin{figure*}
    \centering
    \includegraphics[width=\linewidth]{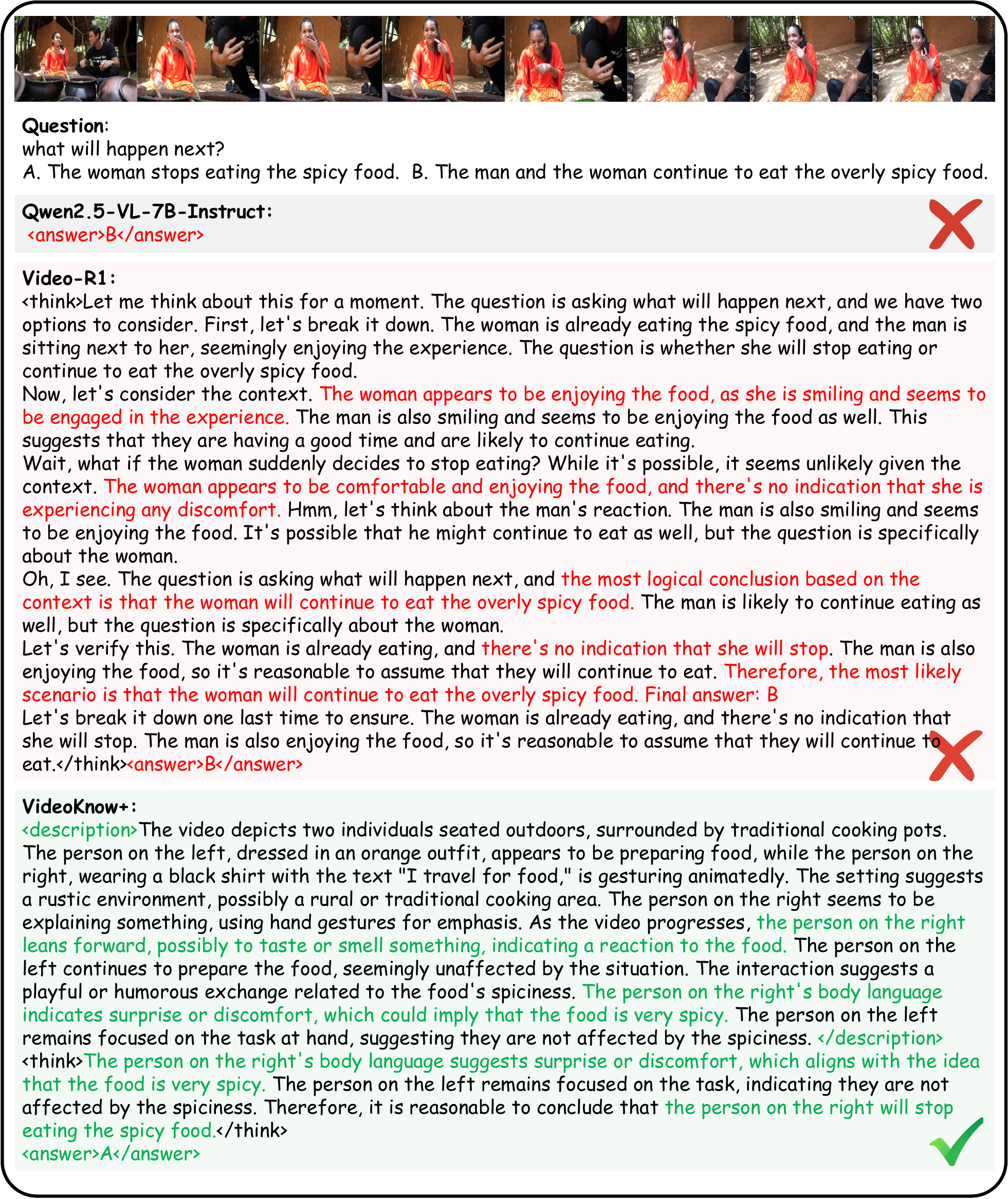}
    \caption{Qualitative comparison on \textit{Event Anticipation} in \BENCH.}
    \label{fig:goodcase5}
\end{figure*}

\begin{figure*}
    \centering
    \includegraphics[width=\linewidth]{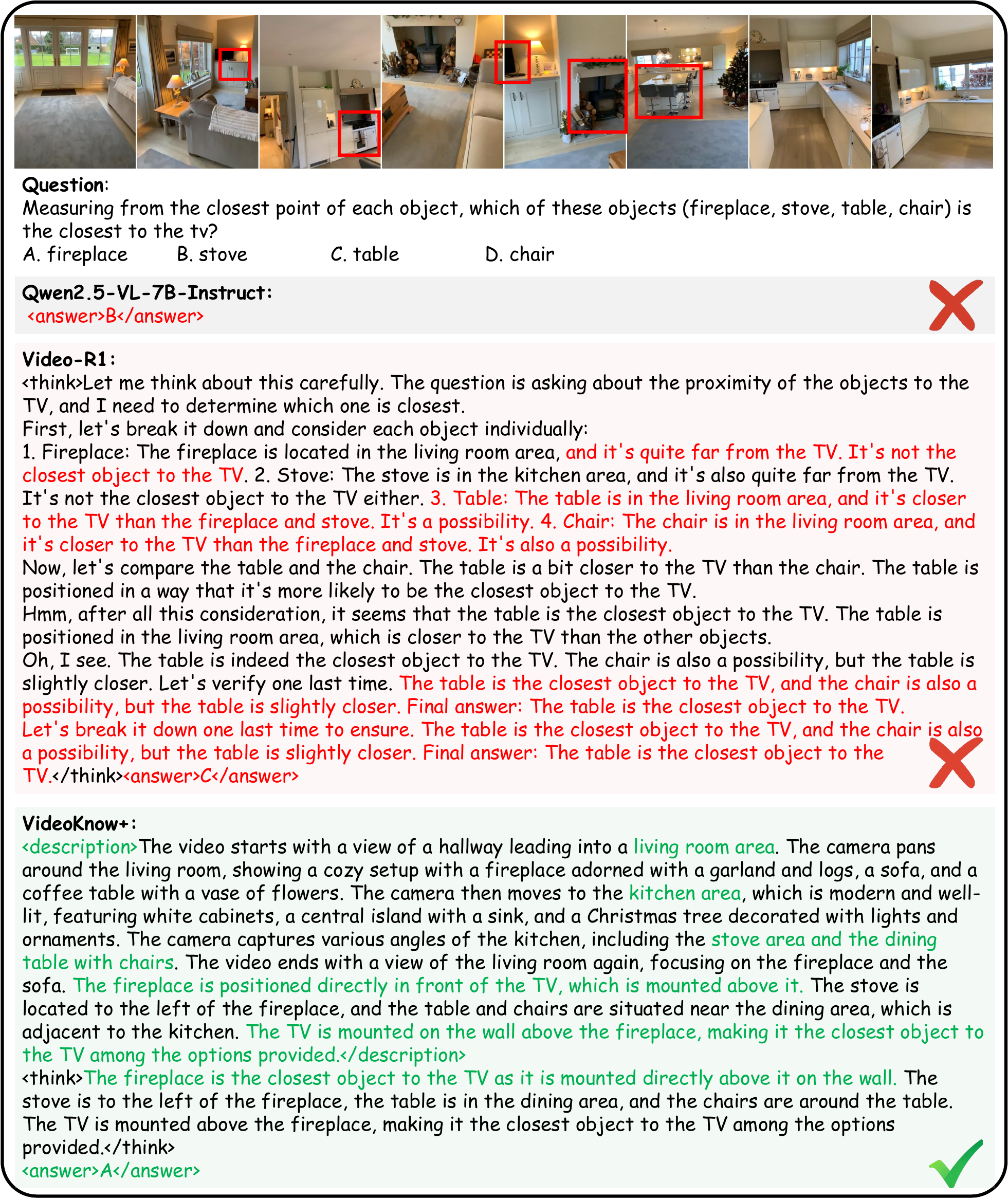}
    \caption{Qualitative comparison on \textit{Spatial Awareness} in \BENCH.}
    \label{fig:goodcase2}
\end{figure*}

\begin{figure*}
    \centering
    \includegraphics[width=\linewidth]{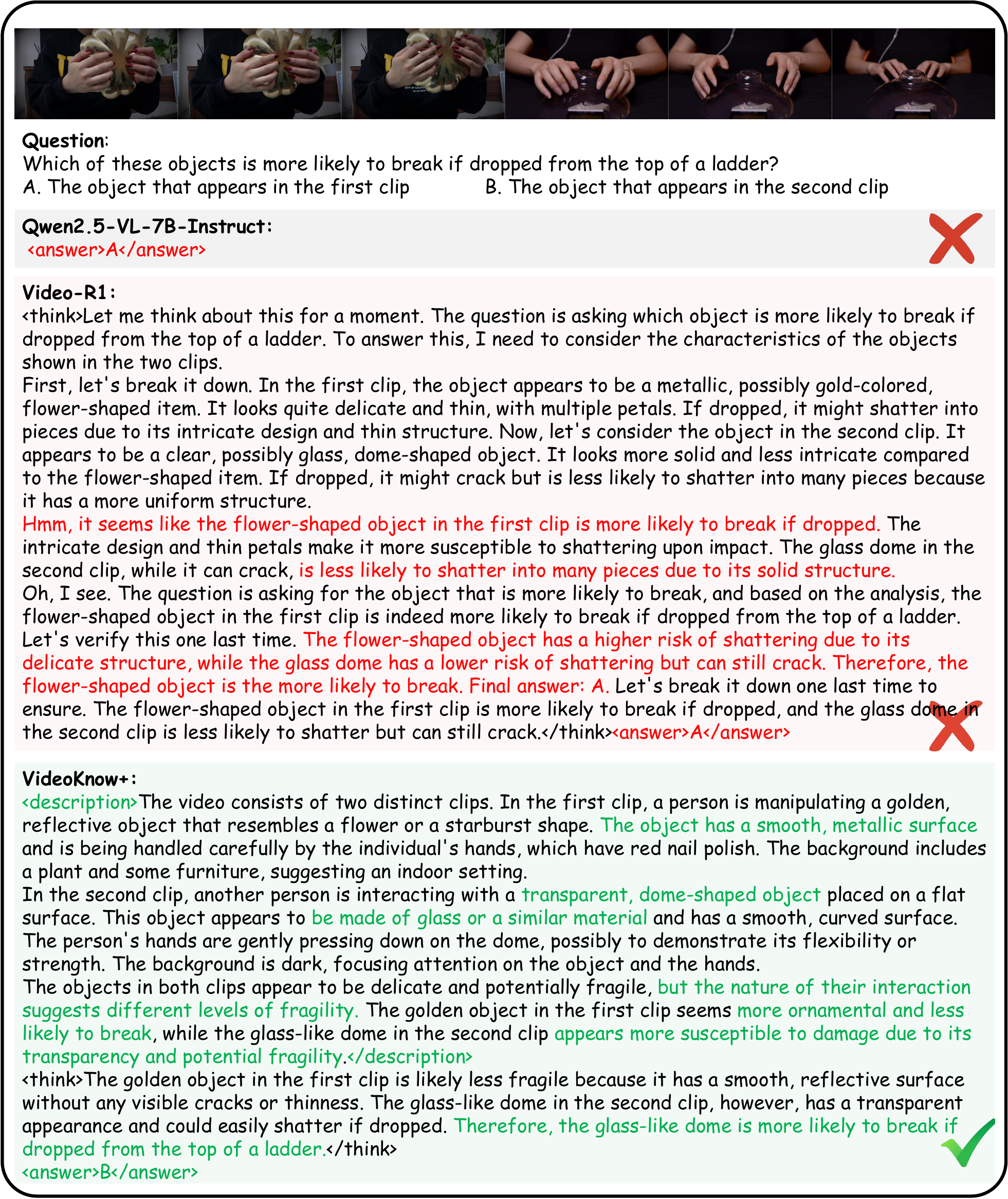}
    \caption{Qualitative comparison on \textit{Object Affordance} in \BENCH.}
    \label{fig:goodcase4}
\end{figure*}

\begin{figure*}
    \centering
    \includegraphics[width=\linewidth]{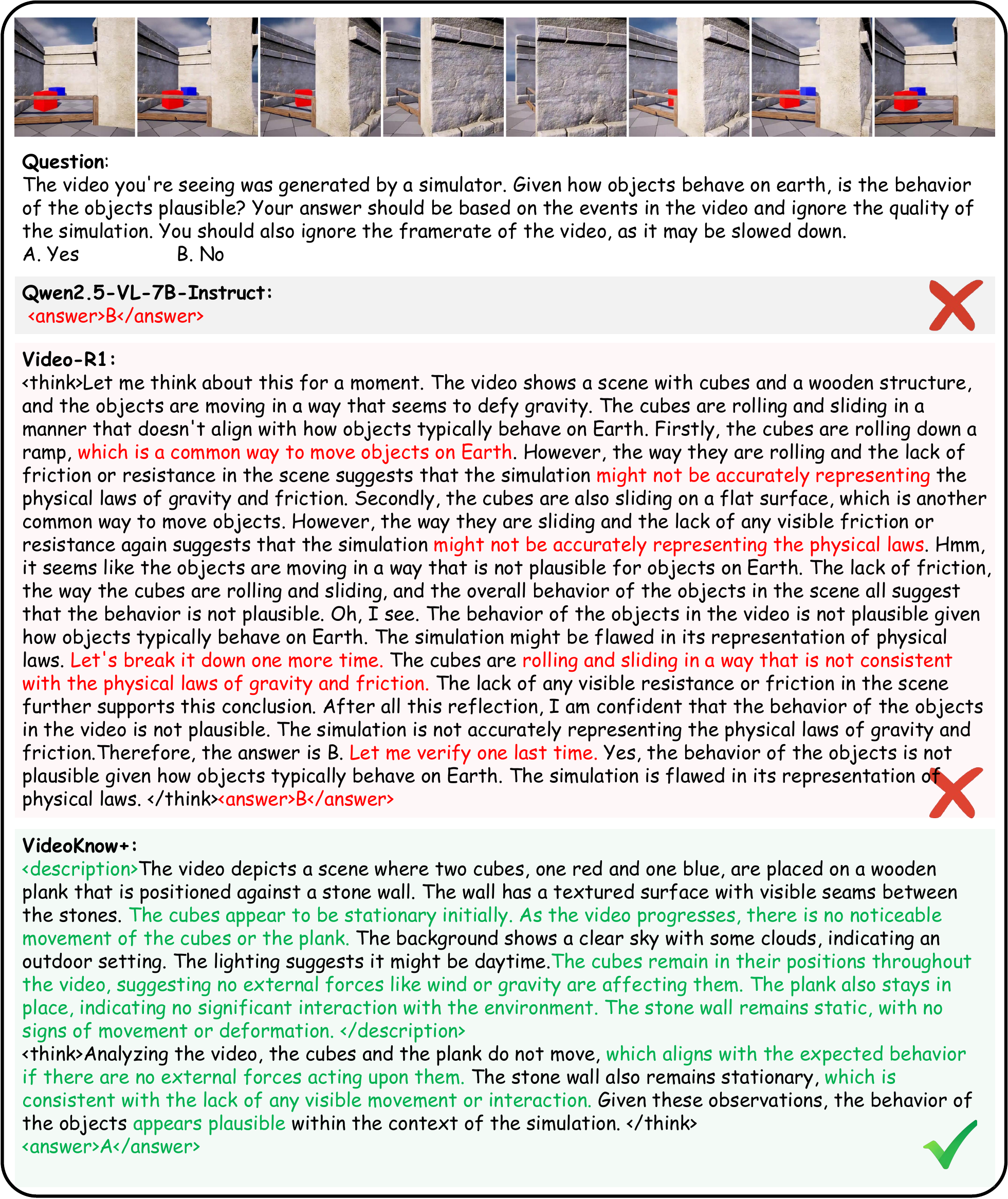}
    \caption{Qualitative comparison on \textit{Intuitive Physics} in \BENCH.}
    \label{fig:goodcase1}
\end{figure*}

\begin{figure*}
    \centering
    \includegraphics[width=\linewidth]{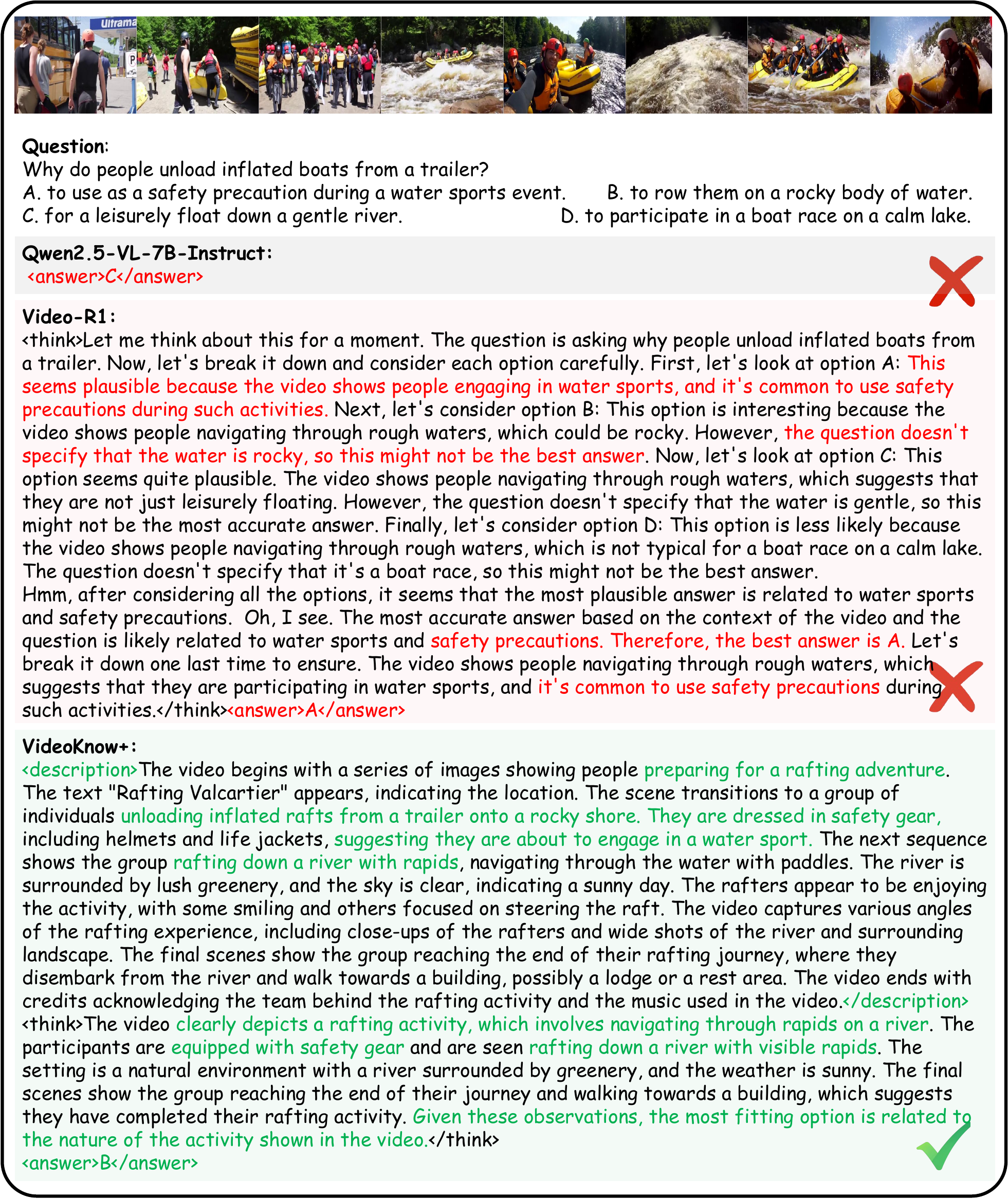}
    \caption{Qualitative comparison on \textit{Subjective Intention} in \BENCH.}
    \label{fig:goodcase3}
\end{figure*}

We compare \MODEL\ with Qwen2.5-VL-7B-Instruct~\cite{bai2025qwen2} and Video-R1~\cite{feng2025video} across diverse tasks in \BENCH. The results in Figs.~\ref{fig:goodcase5}, \ref{fig:goodcase2}, \ref{fig:goodcase4}, \ref{fig:goodcase1}, and~\ref{fig:goodcase3} show that the baselines, whether answering directly or generating a reconstructed chain of thought, still exhibit different forms of hallucination. In contrast, \MODEL\ generates more grounded and reliable outputs, mainly due to the \textit{See--Think--Answer} formulation and the training objective that explicitly emphasizes visual knowledge.

For example, in Fig.~\ref{fig:goodcase5}, Video-R1 produces a lengthy reasoning chain with several rounds of self-reflection but still reaches an incorrect conclusion. This illustrates a limitation of the traditional \textit{Think--Answer} paradigm: textual reasoning can become detached from visual evidence. \MODEL\ instead observes before reasoning, producing concise reasoning grounded in the video content.

We also observe a consistent behavioral difference between the two paradigms. Video-R1 often starts by analyzing answer options rather than first grounding its understanding in the visual input. Although it may later incorporate visual cues, the extracted evidence is often less accurate and less comprehensive than the visual knowledge generated by \MODEL.

\subsection{Shortcomings}
\label{sec:appendix_case_shortcomings}

\begin{figure*}
    \centering
    \includegraphics[width=.95\linewidth]{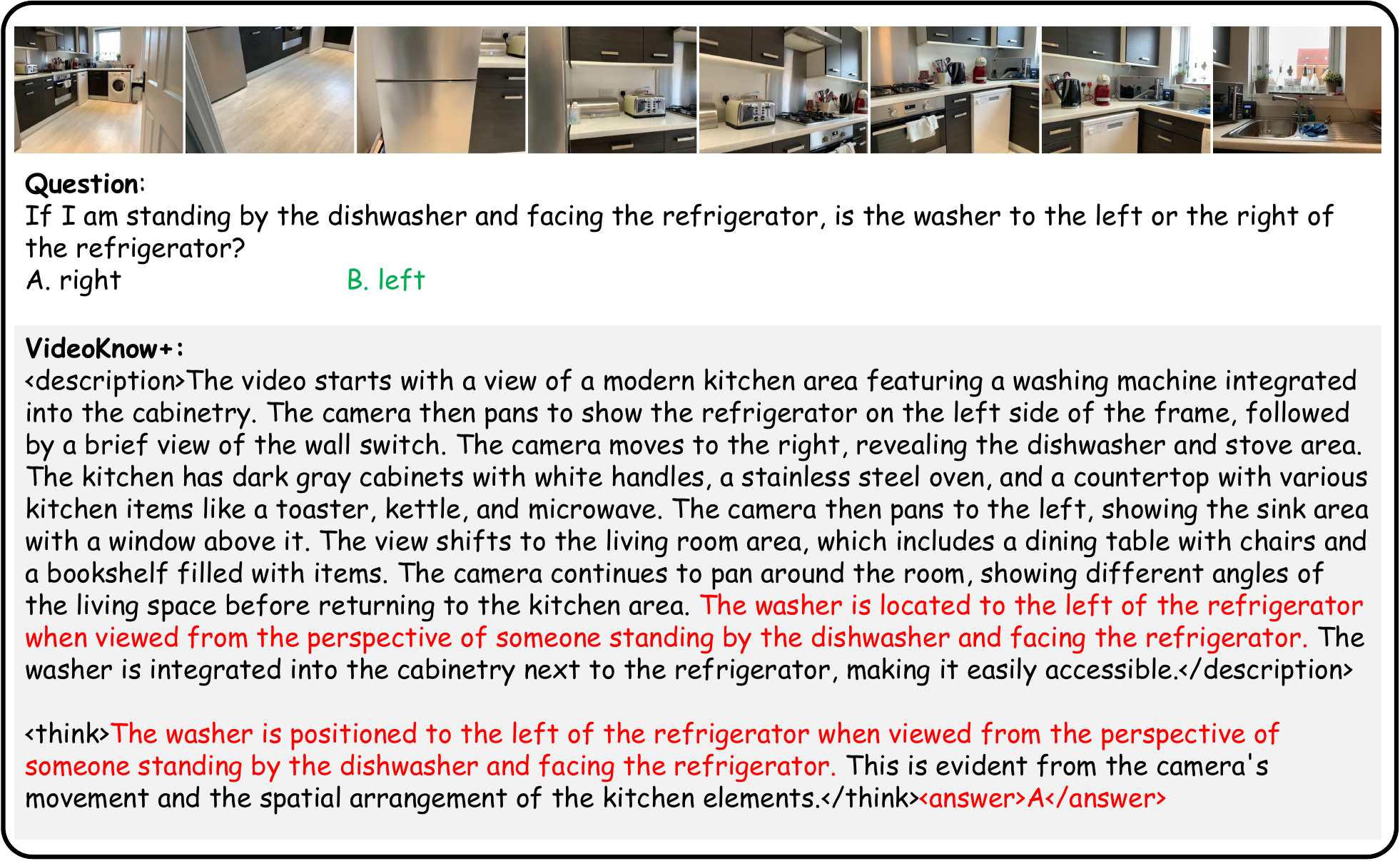}
    \caption{Failure case of \MODEL: inaccurate visual knowledge I.}
    \label{fig:badcase1}
\end{figure*}

\begin{figure*}
    \centering
    \includegraphics[width=.95\linewidth]{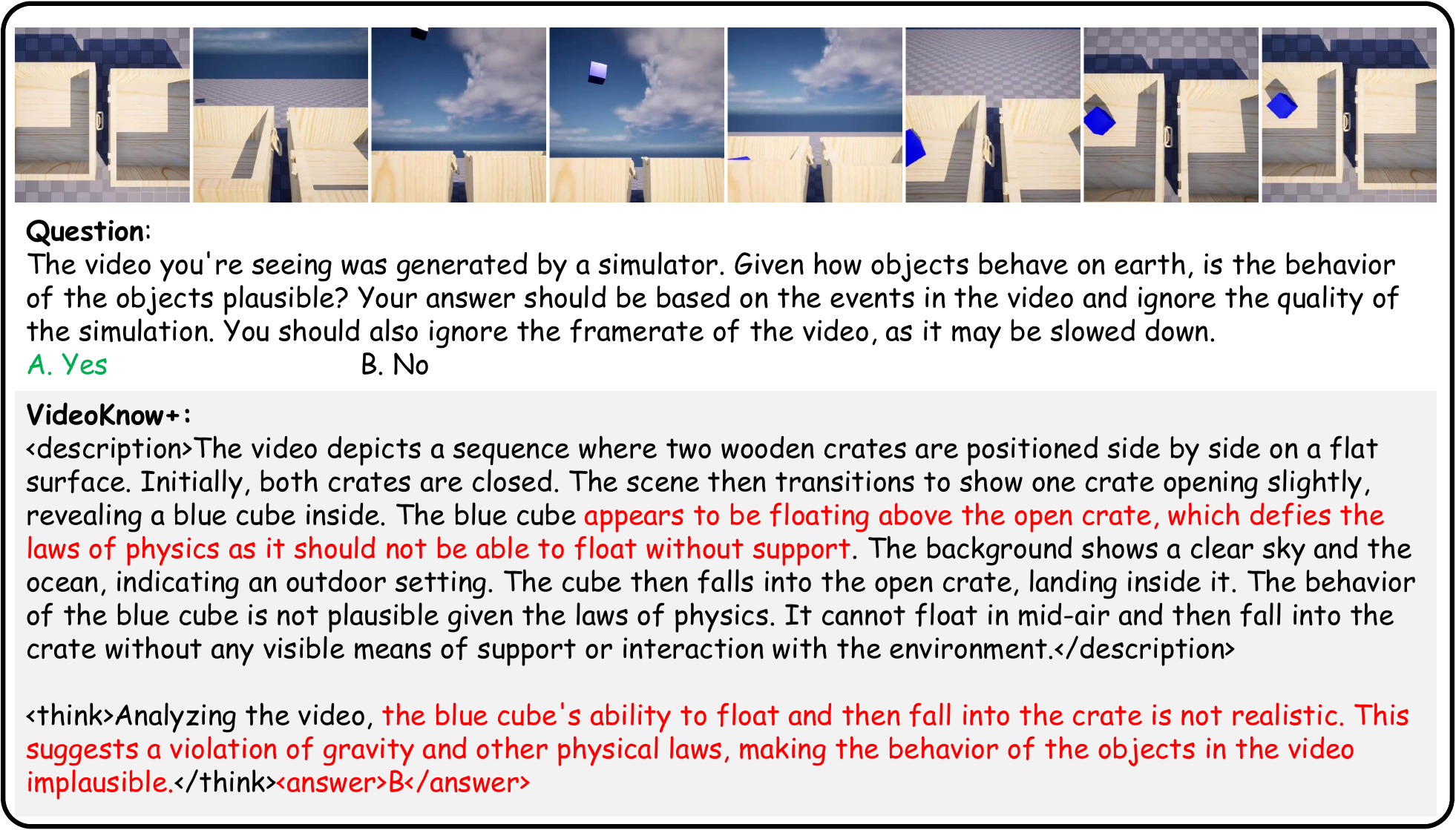}
    \caption{Failure case of \MODEL: inaccurate visual knowledge II.}
    \label{fig:badcase2}
\end{figure*}

\begin{figure*}
    \centering
    \includegraphics[width=.85\linewidth]{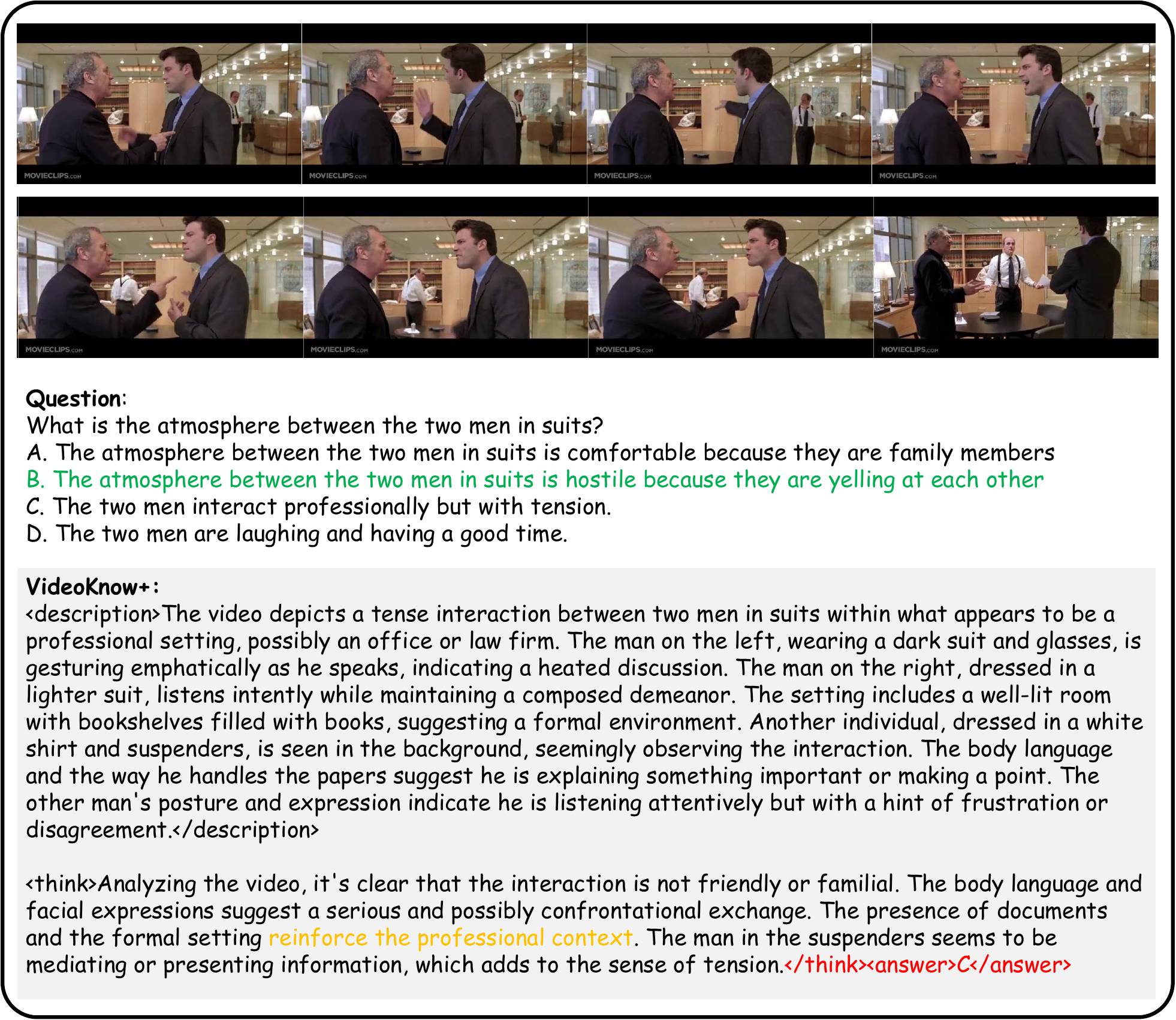}
    \caption{Failure case of \MODEL: insufficient visual knowledge.}
    \label{fig:badcase3}
\end{figure*}

\begin{figure*}
    \centering
    \includegraphics[width=.85\linewidth]{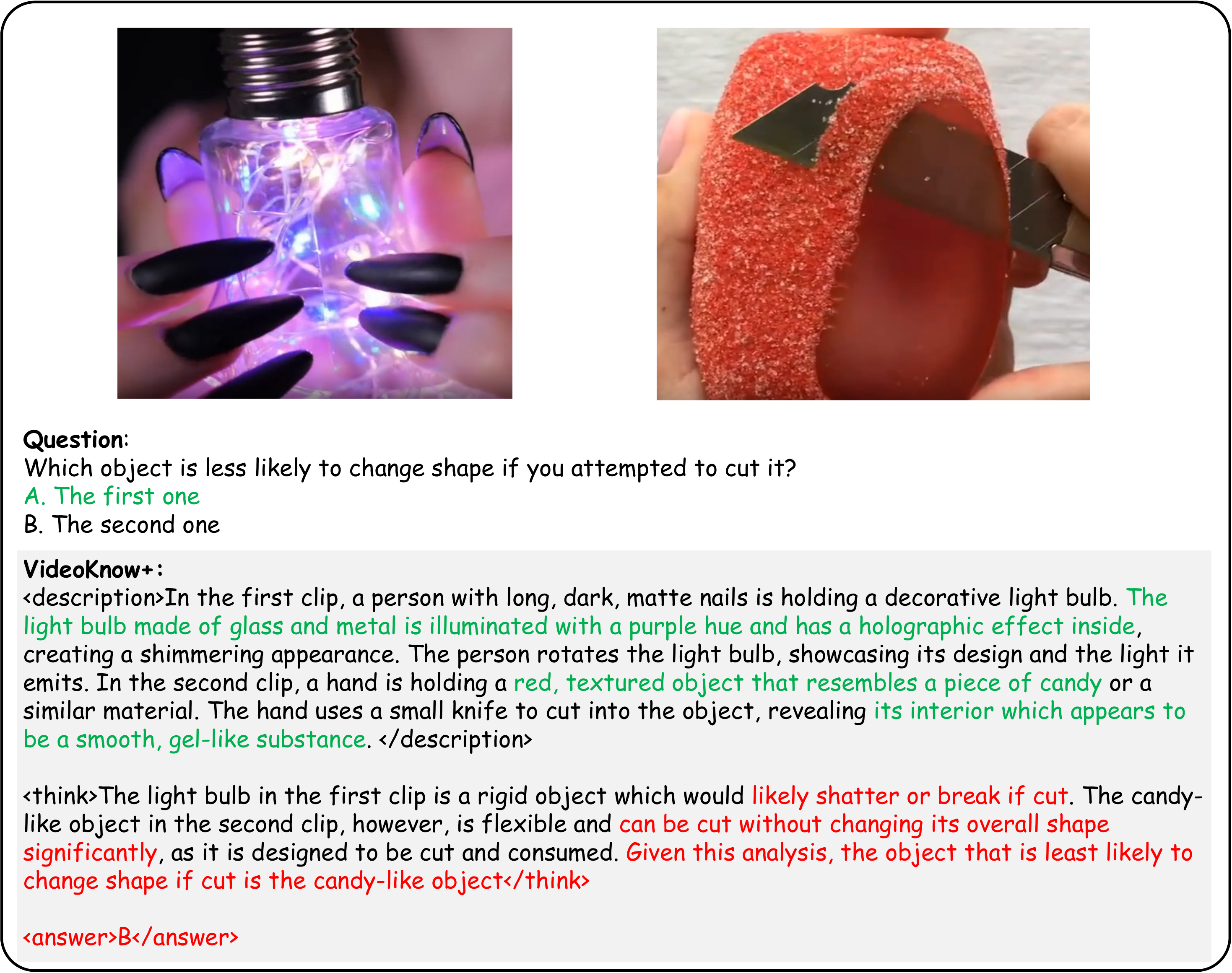}
    \caption{Failure case of \MODEL: insufficient LLM knowledge.}
    \label{fig:badcase4}
\end{figure*}

We further present failure cases where \MODEL\ still encounters difficulty. The errors fall into three categories: (1)~\textbf{inaccurate visual knowledge} (Figs.~\ref{fig:badcase1} and~\ref{fig:badcase2}), (2)~\textbf{insufficient visual knowledge} (Fig.~\ref{fig:badcase3}), and (3)~\textbf{insufficient LLM knowledge} (Fig.~\ref{fig:badcase4}).

In Fig.~\ref{fig:badcase1}, the model inaccurately describes key scene information, especially the relative locations of objects in the room. The following reasoning is therefore anchored on an incorrect visual premise and leads to the wrong answer. A similar issue appears in Fig.~\ref{fig:badcase2}, where the model hallucinates that the blue cube floats above the crate, directly affecting the final prediction.

In contrast, Fig.~\ref{fig:badcase3} reflects missing rather than incorrect visual evidence. Although the generated description is mostly plausible, the model fails to capture the critical cue that two men are yelling at each other. Without this signal, the reasoning drifts toward a generic interpretation of professional tension and misses the hostile atmosphere.

Finally, Fig.~\ref{fig:badcase4} illustrates insufficient LLM knowledge. The model produces a correct visual description and extracts the relevant visual knowledge, but the final prediction is still incorrect due to inadequate language priors in the LLM. These cases show that visual knowledge understanding remains an open challenge even when explicit visual grounding is encouraged.

%% file: tables/traindata_source.tex
\begin{table}[H]
\caption{Data source of \DATA-30K.}
\label{tab:traindata_source}
\centering
\resizebox{.8\columnwidth}{!}{
\begin{tabular}{llc}
\toprule
\textbf{Category} & \textbf{Source} & \textbf{Size (\%)} \\
\midrule
\multirow{3}{*}{General} & LLaVA-Video-178K~\cite{zhang2024video},  & \multirow{3}{*}{12K (40.0\%)}\\ 
&Video-R1-260K~\cite{feng2025video}&\\
&NextQA~\cite{xiao2021next}&\\
\midrule
\multirow{2}{*}{World-Centric} & CLEVRER~\cite{yi2019clevrer}, VSI-100K~\cite{liao2025improved}, & \multirow{2}{*}{9K (30.0\%)} \\
& Intphys~\cite{riochet2018intphys}, STAR~\cite{wu2024star} &\\
\midrule
\multirow{2}{*}{Human-Centric} & EMER~\cite{lian2023explainable}, MAFW~\cite{liu2022mafw}, & \multirow{2}{*}{9K (30.0\%)} \\
& Social-IQ~\cite{zadeh2019social}, CausalVidQA~\cite{li2022representation} & \\
\bottomrule
\end{tabular}
}
\end{table}